\documentclass{article}
\usepackage[a4paper, total={6in, 8in}]{geometry}
\usepackage{graphicx}
\usepackage[utf8]{inputenc}
\usepackage{setspace}
\usepackage{bm}
\usepackage{amssymb}
\usepackage{amsmath}
\usepackage[mathlines]{lineno}
\usepackage{dcolumn}
\usepackage{amsfonts}
\usepackage{bbold}
\usepackage{authblk}
\usepackage{hyperref}
\usepackage{cite}

\title{Using Machine Learning to Anticipate Tipping Points and Extrapolate to Post-Tipping Dynamics of Non-Stationary Dynamical Systems}
\author{Dhruvit Patel \\ {dpp94@umd.edu}} 
\affil{University of Maryland, College Park, Maryland 26742, U.S.A.}
\author{Edward Ott}
\affil{University of Maryland, College Park, Maryland 26742, U.S.A.}

\date{\today}

\begin{document}

\maketitle

\section*{Abstract}

In this paper we consider the machine learning (ML) task of predicting tipping point transitions and long-term post-tipping-point behavior associated with the time evolution of an unknown (or partially unknown), non-stationary, potentially noisy and chaotic, dynamical system. We focus on the particularly challenging situation where the past dynamical state time series that is available for ML training predominantly lies in a restricted region of the state space, while the behavior to be predicted evolves on a larger state space set not fully observed by the ML model during training. In this situation, it is required that the ML prediction system have the ability to extrapolate to different dynamics past that which is observed during training. We investigate the extent to which ML methods are capable of accomplishing useful results for this task, as well as conditions under which they fail. In general, we found that the ML methods were surprisingly effective even in situations that were extremely challenging, but do (as one would expect) fail when ``too much" extrapolation is required. For the latter case, we investigate the effectiveness of combining the ML approach with conventional modeling based on scientific knowledge, thus forming a hybrid prediction system which we find can enable useful prediction even when its ML-based and knowledge-based components fail when acting alone. We also found that achieving useful results may require using very carefully selected ML hyperparameters and we propose a hyperparameter optimization strategy to address this problem. The main conclusion of this paper is that ML-based approaches are promising tools for predicting the behavior of non-stationary dynamical systems even in the case where the future evolution (perhaps due to the crossing of a tipping point) includes dynamics on a set outside of that explored by the training data.

\section{Introduction}

Predicting the time evolution of a dynamical system is a problem at the heart of many fields. While some prediction problems focus on forecasting the evolution of the values of a set of system state observables over a timescale that is on the order of the characteristic time ($\tau_s$) on which observables vary, in other situations one is concerned with predicting the statistical properties of observables over a timescale ($\tau_{ns}$) that is much longer than the time over which the details of system state variation can be usefully forecasted ($\tau_{ns} \gg \tau_s$). Examples of the former include predicting daily rainfall, wind speed, and temperature (i.e., weather forecasting), while examples of the latter include predicting variations of average patterns associated with rainfall, wind and temperature over years. (Motivated by the terminology in atmospheric science, we use the term ``climate" to refer to the long-term characteristics of typical orbits of any dynamical system.) Prediction is often particularly difficult since many systems of interest, such as the terrestrial climate system, can be highly complicated, and knowledge of some of their scientific principles, parameters, and boundary conditions may be incomplete, inaccurate, or unknown. In addition, such systems may have a wide range of spatial and temporal scales that cannot be resolved using conventional numerical methods. The long term statistics of the dynamics of such systems is often heuristically modeled using noisy, \emph{non-stationary} dynamical systems which themselves may depend on a set of time-dependent parameters. It is well-known that for different \emph{fixed} parameter settings, a noiseless \emph{stationary} dynamical system (i.e., a system with no explicit time-dependence of the system itself) can, depending on its (time-independent) parameters, exhibit a variety of behaviors ranging from periodic to chaotic. We refer to qualitative changes of the attractor orbits of \emph{stationary} systems occurring with variation of (time-independent) system parameters as \emph{bifurcations}. In the case where the system is \emph{non-stationary}, e.g., due to time-dependent parameters, the basic system dynamics may change with time. Moreover, if, for example, the time-dependent system temporally drifts through a critical parameter region, the state evolution of the non-stationary system can experience rapid change whereby the statistical behavior of its dynamics is qualitatively and quantitatively altered \cite{tel_tipping,feudel}. We refer to such changes in the dynamics of \emph{non-stationary} systems as ``tipping points".

Machine learning has been widely applied to the problem of determining both the short-term future state evolution \cite{jaeger_science,pathak_prl,pathak_chaos1,vaughan,griffith,shen,follmann,herzog,chatto,chen,huang,fan,faqih,gauthier,borra} and the long-term ``climate" \cite{pathak_chaos2,lu_chaos,antonik,nguyen} of \emph{stationary} dynamical systems. In this paper, we use the term ``climate" to denote long-term statistical properties of the evolution of a dynamical system. For stationary systems with ergodic dynamics, this includes, for example, obtaining the distribution of states, Lyapunov exponents, temporal correlation functions, Fourier power spectra, etc. associated with typical long trajectories of the system. However, for non-stationary systems the estimation of local-in-time state distributions, Lyapunov exponents, Fourier power spectra, and temporal correlation functions can often be problematic. We note, however, that a more well-defined generalization of a time varying system state distribution for non-stationary systems is available through the concept of ``snapshot attractors" (also called "pullback attractors"), and we shall make use of this. In particular, a snapshot attractor, at any given time $t$ is obtained by considering an ensemble of states obtained from an ensemble of trajectories initiated from many randomly chosen initial conditions in the far past. See Refs. \cite{ruelle,drotos1,drotos2,romeiras,chekroun,patel} for a more detailed discussion. While prediction of stationary systems by machine learning (ML) has received much recent attention, less progress has been made in applying ML to the problem of predicting the time evolution of \emph{non-stationary} dynamical systems, particularly of their climate and of the tipping points they may experience. Refer to \cite{chen,lim,xing,shi} for recent works which apply machine learning to the problem of predicting the short-term state evolution of non-stationary dynamical systems and to \cite{patel, pershin,bury,kong,xiao} for recent works which aim to address the problem of predicting changing statistical properties of non-stationary dynamical systems (including anticipating tipping points). In previous work \cite{patel} we demonstrated that ML provides a promising avenue for predicting the climate of a non-stationary dynamical system using the time series of its past states and knowledge of a non-stationarity-inducing system parameter time dependence. It was shown that a machine learning model can anticipate tipping points in a non-stationary dynamical system and, in some cases, predict post-tipping-point dynamics which are fundamentally different from those it was trained on. 

The goal of this work is to further develop, devise, and test ML techniques for the prediction of non-stationary dynamical systems that undergo a tipping point transition. A main focus of our work is on situations where \emph{the observed pre-transition motion is constrained to a smaller restricted subset of the state space region than that on which the post-transition motion evolves}. 

In contrast, in previous work on predicting tipping points and the associated post-tipping-point dynamics \cite{patel,kong,xiao} ML prediction was considered for cases in which the training data was obtained from orbits that typically explored large state space regions that included all or most of the state space visited by the predicted future orbits. Thus, in this prior work the ML predictor was directly aware of dynamical system information needed for the prediction of the future behavior, e.g.,  after a predicted tipping point. In some other previous works only the occurrence of a tipping point, but not the post-tipping-point behavior, was predicted. These latter works anticipate the occurrence of a tipping point based on observation of a pre-tipping-point orbit subject to dynamical noise, and use the fact that, as the tipping point is approached, the effect of the noise on the orbit increases. For example, in one work of this type \cite{bury} a deep learning technique was developed by training on a library of mathematical models to recognize dynamical response to noise that characterizes a system as it approaches a tipping point. Although quite useful, such techniques yield no information about the post-tipping-point dynamics of the system. In contrast to the above two different cases, in this paper we consider the situation where the ML predictor, although trained on a pre-tipping-point system trajectory which evolves on a smaller state space set contained within the larger set explored by the future post-tipping-point trajectory, is able to anticipate the tipping-point transition and \emph{extrapolate} its learning from the neighborhood of the pre-tipping-point training data into the larger regions of the system state space not explored, or only sparsely explored, in the training data to predict the post-tipping-point dynamics.

A further contribution of our paper relative to previous related papers using ML for prediction of non-stationary behavior \cite{patel,kong,xiao} is that those previous works considered the case where the system non-stationarity was induced by time variation of a parameter of an otherwise unknown system, and this parameter time variation was assumed to be known and was used as an input to the ML prediction system. In this current paper, on the other hand, we consider the case where knowledge of the type described above may not be available. 

Furthermore, motivated by our finding that using the standard hyperparameter optimization validation scheme for this type of forecasting did not consistently yield useful results in our test cases, we accordingly \emph{introduce a new hyperparameter optimization strategy (which we use in the numerical experiments of this paper) for this purpose}.  We believe that this hyperparameter determination scheme may be generally beneficial for ML prediction of non-stationary systems.

Tipping point transitions between different dynamical states where the pre-transition state dynamics is constrained to a smaller set of the system state space, or to a different region of the system state space, than the post-transition orbits are observed in a wide array of natural systems \cite{tel_tipping,feudel}. Such transitions are commonly found in various terrestrial climate models \cite{ghil_review,dekker,tantet}, ecosystem models \cite{zelnik,amemiya,fussmann}, epidemics \cite{alonso,muk}, and physical and engineering applications (e.g., intermittency \cite{pomeau} and crisis \cite{grebogi} transitions in plasmas \cite{chiriac,cheung,cassak,chian}, lasers \cite{manffra,sanmartin}, electrical and power systems \cite{dobson,kim,huang2,guerrero}, thermoacoustic systems \cite{penelet,guan,kabiraj}, hydrodynamical systems \cite{ringuet,suresha}, and electrochemical systems \cite{krischer}). Thus, developing methods to predict the climate evolution of non-stationary dynamical systems which may tip into a state characterized by motion of the system that may visit previously unexplored, or only sparsely explored, regions of its state space is a problem of very general importance. We emphasize, however, that, as is the case in general for methods employing extrapolations, our method has limitations. In particular, extrapolations are more likely to fail as the "amount" of the attempted extrapolation from the known case increases. A case illustrating this point is given in Sec. 3.1 where, when considering a situation where the non-stationary drift of the system is too slow, predictive extrapolation of the system behavior fails, but is then found to succeed when dynamical noise is assumed to influence the system evolution. (Evidently, the beneficial role of the dynamical noise in this case is to increase the state space set that is sampled by the training data.) A related example is given in Sec. 3.4 where we begin by reporting a failed attempt to extrapolate and predict behavior through a tipping-point associated with a subcritical (hysteretic) Hopf bifurcation using a purely data-driven ML model. In order to enable prediction in such a case, we then consider a prediction system scheme in which an ML model is combined with a knowledge-based model component. Although the knowledge-based component in the example considered is so inaccurate that it cannot make useful predictions on its own, we show that using it in a combined ML-based/knowledge-based hybrid prediction system enables good predictions of a tipping point transition as well as of the post-tipping-point behavior. Based on the discussion of Sec. 3.4, we hypothesize that the ML prediction of post-tipping climate dynamics purely from pre-tipping training data will usually only be possible for tipping point processes mediated by non-hysteretic stationary system bifurcations.

In what follows, we use reservoir computing \cite{luk,jaeger2,maass} as the ML platform. Reservoir computing has previously been successfully used to predict the time evolution of dynamical systems \cite{jaeger_science,pathak_prl,pathak_chaos1,griffith,chatto,pathak_chaos2,lu_chaos,antonik,patel}, and its training is computationally inexpensive since it can be accomplished via a simple linear regression. This allows us to rapidly test different methodologies and various test system scenarios. We expect other types of machine learning, such as deep learning, to also work well using the methods presented in this paper. 

The rest of the paper is structured as follows. Section II presents a brief introduction to the reservoir computing setup and training, as well as to our hyperparameter optimization scheme for non-stationary systems. In Section III we numerically demonstrate the use of ML to anticipate tipping point transitions and post-tipping behavior from motion in a restricted state space region to motion that explores substantial state space regions not previously visited. The example test systems used in Section III for generating the training data are the three-dimensional Lorenz '63 system \cite{lorenz} (Sec. 3.1), the Ikeda map \cite{hammel} (Sec. 3.2), and spatiotemporally chaotic Kuramoto-Sivashinsky partial differential equation \cite{kuramoto,siva} (Sec. 3.3). In Section IV we conclude with a brief summary of our main points. 

\section{Reservoir Computing Background}

\subsection{Setup and Training}

Reservoir computing (see the review paper Ref. \cite{luk} for details) is a framework for efficiently training recurrent neural networks. In the implementation we use here, it consists of three major components: (1) a fixed input layer, (2) a fixed reservoir (in our case, a network of neuron-like nodes with recurrent connections), and (3) a \emph{trainable} output layer. In this paper, we employ a setup similar to that used in Ref. \cite{patel} for predicting noisy non-stationary dynamical systems (see Sections II, IV-A, and IV-B of Ref. \cite{patel}). We briefly review the setup and the training/prediction procedure below. 

We assume the availability of measured data from some time in the past, $t=-T$, to the present, $t=0$, which consists of a set of $L$ measured state variables from a dynamical system of interest, represented as a $L$-dimensional vector $\bm{v}(t) = [v_1(t),v_2(t),...,v_L(t)]^T$, which we desire to predict for $t>0$, plus a $S$-dimensional vector $\bm{s}(t)$ of additional variables representing any other available information that may aid in the prediction of $\bm{v}(t)$. During training ($-t_t \leq t \leq 0$ for some training length $t_t \leq T$, where $(T-t_t)$ is assumed to be long enough that transient behavior, associated with start-up, has decayed away), the input to our reservoir computer system will be the vector $\bm{u}(t)=[\bm{v}(t); \bm{s}(t)]^T$, whose dimension we denote by $K=L+S$. Our objective is to predict $\bm{v}(t)$ for $t>0$. If we use a reservoir of $N$ nodes, then the input layer will be a ($N \times K$)-dimensional matrix denoted $W_{in}$. We choose to construct this matrix by randomly selecting one element of each row to be a nonzero number randomly chosen from a uniform distribution on the interval $[-\chi,\chi]$. The reservoir state is denoted by a vector $\bm{r}(t) = [r_1(t), r_2(t), ..., r_N(t)]^T$, where $r_i(t)$ is the scalar state of the $i^{th}$ reservoir node. The reservoir state evolves dynamically in time according to 

\begin{equation}
    \bm{r}(t+\Delta t) = (1-\alpha)\bm{r}(t) + \alpha \tanh(A\bm{r}(t) + W_{in}\bm{u}(t) + b_r\mathbb{1}_{N\times 1}),
    \label{eqn:res_eq}
\end{equation}

where $A$ is the $N \times N$ reservoir network adjacency matrix, $\alpha$ is the ``leakage parameter", and $b_r$ is a constant bias. The reservoir adjacency matrix $A$ is constructed as a directed random Erdos-Renyi graph of $N$ nodes, and degree $\langle d \rangle$, where both $N$ and $\langle d \rangle$ are ``hyperparameters". The output layer is a $(L \times (N+K+1))$ matrix of trainable weights, $W_{out}$. The matrix $W_{out}$ is chosen (``trained") by minimizing the squared Euclidean distance between the target states, $\bm{v}(t)$ for $t<0$, and the projection of the high-dimensional reservoir states, the input vector, and a constant bias onto a space of dimension $L$, $W_{out}[\bm{r}(t);\bm{u}(t-\Delta t);1]$. This is a linear regression problem over the observed data and can be solved by minimizing the following ``cost function", 

\begin{equation}
    \frac{1}{t_t} \sum_{-t_t\leq t \leq 0} ||W_{out}[\bm{r}(t);\bm{u}(t-\Delta t);1] - \bm{v}(t) ||^2 + \lambda||W_{out}||^2,
    \label{eqn:res_lsq}
\end{equation}

where $\lambda$ is the Tikhonov regularization parameter, the term $\lambda ||W_{out}||^2$ is added to prevent over-fitting, and $||W_{out}||^2$ is the Frobenius norm of $W_{out}$. Once $W_{out}$ is obtained, single-step prediction of $\bm{v}(t + \Delta t)$ is simply given by $\bm{v}(t + \Delta t) \approx \tilde{\bm{v}}(t + \Delta t) = W_{out}[\bm{r}(t + \Delta t);\bm{u}(t);1]$.

Following the method of Sec. IV-B in Ref \cite{patel} for the multi-step prediction of non-stationary dynamical systems, we take $K=L+1$ with $s(t)$ chosen to be the scalar quantity $at + b$, for $a$ and $b$ being constants ($S = 1$). Thus $s(t)$ may be considered a linear control signal (instead of the non-stationarity-inducing bifurcation parameter of the system of interest as in Sec IV-A in Ref \cite{patel}. The linear control signal $s(t)$ is provided to the reservoir at every step of the training and prediction process. Multi-step prediction is then obtained by using the single-step prediction $\bm{\tilde{v}}$ at time $t$ as the input for prediction at the next step at time $t + \Delta t$ (i.e., we set $\bm{u}(t+\Delta t) = [\tilde{\bm{v}}(t+\Delta t);s(t+\Delta t)]$). The linear control signal allows the reservoir to act as time-dependent mapping of input $\bm{v}(t)$ to output $\tilde{\bm{v}}(t+\Delta t)$.

In many application settings the observed orbits are influenced by \emph{dynamical} noise. Since the reservoir computer learns the input-to-output mapping via linear regression, it is possible (as demonstrated in Refs. \cite{patel,chandra}) that given the time-series of a noisy dynamical system, the reservoir computer learns a mapping which closely approximates the underlying noiseless system. This may be undesirable since dynamical noise is present in the real system whose climate we wish to predict, and it can greatly influence the dynamics (e.g., it may induce intermittent bursting behavior near a bifurcation). Since we would like to capture these noise-related climate effects in our predictions, we will use a setup described in detail in Sec. IV-B of Ref \cite{patel} in which the reservoir system is modified to produce stochastic outputs mimicking the effect of dynamical noise present in the measured variables that are predicted. In particular, upon training the reservoir computer on the time series in the manner described above, we calculate the difference between the noisy target trajectory used for training, $\bm{v}(t)$, and the corresponding trajectory reconstructed from the reservoir states during training with the learned $W_{out}$ matrix, $W_{out}[\bm{r}(t);\bm{u}(t-\Delta t);1]$. The point-wise errors, \{$\bm{v}(t)-W_{out}[\bm{r}(t);\bm{u}(t-\Delta t);1]$\} for $t\leq0$, between the reservoir-reconstructed training trajectory and the true trajectory can then be treated as a distribution which approximates the dynamical noise distribution forcing the true system dynamics (assuming that the reservoir size is large enough that the reservoir computer's learned approximation to the underlying noiseless dynamics of the true system \cite{chandra} is good). Then, during prediction, we randomly sample from this error distribution and add that to the one-step prediction at each step (and this perturbed one-step prediction is then fed back in as the input at the next time step). This scheme (see Ref. \cite{patel}) will be used for all numerical experiments in this paper.

\subsection{Choosing Hyperparameters for Prediction of Non-Stationary Systems}

Next, we turn our attention to the problem of selecting appropriate ``hyperparameters", i.e., parameters of the machine learning model which are not learned but instead are chosen a priori. For reservoir computing in our case, such parameters include the number of reservoir nodes ($N$), the degree ($\langle d \rangle$) and spectral radius ($\rho_s$) of the reservoir adjacency matrix, the strength of the input-to-reservoir coupling ($\chi$), the reservoir leakage term ($\alpha$), the Tikhonov regularization parameter ($\lambda$), the reservoir activation bias ($b_r$), the slope ($a$) and intercept ($b$) parameters of the linear control signal, the strength of the observational noise added to the training data ($\epsilon_0$) and the number of passes of the training data during training (see \cite{wikner} and Sec. VI of Ref \cite{patel} regarding the addition of observational noise and multiple passes of training data during training to aid in the stability of machine learning predictions), and the length of the training data ($t_t$). 

\begin{figure}[h]
    \centering
    \includegraphics[scale=0.5]{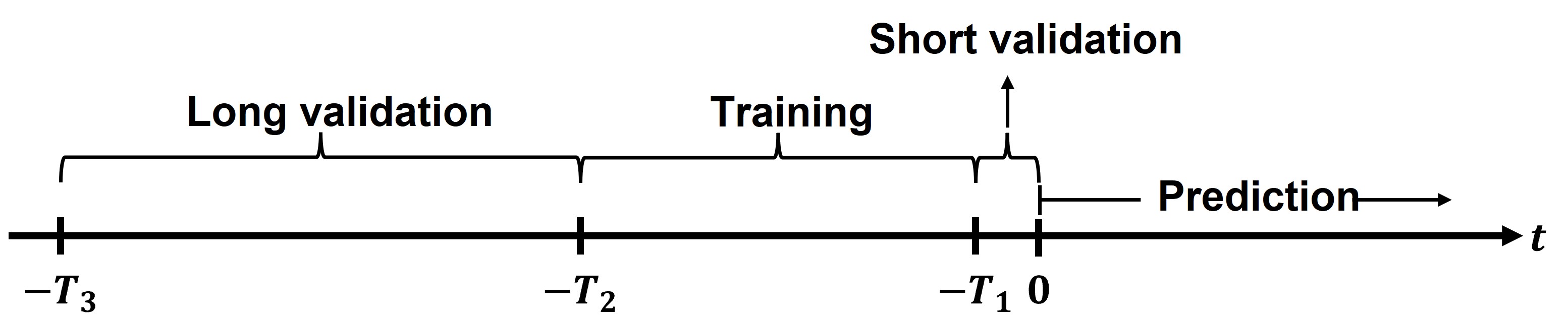}
    \caption{Partition of the training data into the long validation set ($t \in [-T_3,-T_2)$), training set ($t \in [-T_2,-T_1]$), and short validation set ($t\in(-T_1,0]$) for the proposed hyperparameter optimization scheme.}
    \label{fig:hp_optim}
\end{figure}

We choose an appropriate set of hyperparameters by performing a grid-search optimization in which, for each set of hyperparameters, we train the reservoir computer on a training data set and then evaluate its performance on a validation data set (which is disjoint from the training data set). We would like the predictions to (1) produce good multi-step prediction of the state of the system and (2) produce good climate forecasts (i.e., even after short-term prediction of the state of the system has failed, the predicted orbit continues to exhibit dynamics similar to those of the true system). Hence, we choose a criterion for evaluating a trained reservoir's performance on a validation set to capture both of these qualities. We do this in the following way. First we partition the available $t<0$ time series data string into $3$ parts: (1) a short validation set, (2) a training set, and (3) a long validation set. The short validation set will be the data corresponding to $-T_1 < t \leq 0$, where $T_1$ is on the order of a few times (say $20$) $\tau_s$, the characteristic time over which the state of the system varies, e.g., if the Fourier series of a state variables has a pronounced peak, we might choose $\tau_s$ to correspond with this peak, alternatively we might choose $\tau_s$ to be $1 / \lambda_L$ for a chaotic system where $\lambda_L$ is the largest positive Lyapunov exponent of the system of interest. The training data set will correspond to $-T_2 \leq t \leq -T_1$ where the training length $t_t = T_2 - T_1$ is a hyperparameter. The long validation data set will correspond to $-T_3 \leq t < -T_2$, where $(T_3 - T_2)$ will be on the order of many times $\tau_s$ (say $200\tau_s$). See Fig. \ref{fig:hp_optim}. After training the output weights of the reservoir computer on the training data, we first determine how well this trained reservoir unit can predict the short-term state of the true system by considering the median ``valid time" over a set of $m_1$ predictions (as described below). The valid time of a predicted trajectory is obtained by calculating a normalized Euclidean error between the reservoir-computer-predicted trajectory and the true system trajectory and monitoring where this error crosses a chosen threshold (as described in \cite{pathak_chaos1} Sec. III-C). To calculate the median valid time, we first randomly select $m_1$ starting points in the short validation set near $t = -T_1$, resynchronize the reservoir computer to a short segment of data (say of length $\tau_s$) starting at each of the randomly selected starting points (i.e., run it in ``open-loop") and then predict (``close the loop"). The median of the $m_1$ valid times is then called the median valid time for the trained reservoir computer used (for that specific set of hyperparameters). The stability of predictions can be assessed by similarly making $m_2$ independent predictions over the long validation set, starting near $t = -T_3$. In this case, instead of calculating the valid time of each trajectory, we calculate the Wasserstein distance [see the beginning of Sec. III (e.g., Eq. (\ref{eqn:gamma})) and \cite{vallender}] between the distribution of states of the predicted trajectory and that of the true system.  For multivariate systems, we simply compute the Wasserstein distance for each variable and then average them. (In practice, we calculate the Wasserstein distance by using the range of the true system orbit as the integration bounds for the Wasserstein integral). The average Wasserstein distance for each set of hyperparameters is then obtained by averaging this quantity over the $m_2$ predictions. For each set of hyperparameters, we calculate the median valid time ($t_v$) and the average Wasserstein distance ($E_W$). Then we calculate the following quantity for each hyperparameter set

\begin{equation}
    \mathcal{E} = (E_W / \mathrm{max}(E_W)) - (t_v / \mathrm{max}(t_v)),
    \label{eqn:hp_optim}
\end{equation}

where the $\mathrm{max}( )$ of $E_W$ and $t_v$ is calculated over all hyperparameter sets. Finally, the set of hyperparameters which minimize $\mathcal{E}$ are taken as the "appropriate" set of hyperparameters. This is the procedure which was followed in all numerical experiments performed in this paper.

We now discuss the intuition behind this procedure. We would like to obtain hyperparameters which allow the ML to (1) accurately learn the target non-stationary system dynamics near $t=0$ (since we would like to predict for $t>0$), and (2) produce predictions which capture good climate statistics for periods of length much greater than $\tau_s$. The first of these is assessed by evaluating the median valid time. The second is assessed by evaluating the average Wasserstein distance over the long validation set. Furthermore, prediction on the long validation set allows us to avoid choosing hyperparameter sets which yield an ML model that predicts orbits that become unstable when operated in the closed-loop prediction phase for long periods of time, even if the ML model produces good short-term forecasts. To ensure that our choice of hyperparameters allows the ML to accomplish the above two goals, we combine the median valid time and average Wasserstein distance metrics as done in Eq. (\ref{eqn:hp_optim}). [A possible generalization of Eq. (\ref{eqn:hp_optim}) is to weigh the two terms unequally, but in our numerical tests, for the particular systems we considered we found that equal weighing (as in Eq. (\ref{eqn:hp_optim})) was sufficient.] The temporal ordering (Fig. \ref{fig:hp_optim}) of the training data set and the short and long validation sets are chosen so that we perform training and the median valid time tests closest to the present time, near $t=0$, so as to most accurately capture dynamics which are most relevant for prediction. We found that, in many cases, this ordering scheme was essential for obtaining good results. In addition, we note that, even when long-term stability of the reservoir-computer-predictions was not an issue, incorporating the long validation set into the hyperparameter optimization in the above way resulted in more accurate prediction of tipping point transitions and post-tipping-point transition dynamics (see Appendix A for an example comparison).

In all of our numerical experiments, all components of the reservoir input vector $\bm{v}$ are normalized by their root-mean-square values (taken over the observed data for $t<0$). In addition, we set $s = a k + b$ where $k=1,2,3,...$ denotes the reservoir time step. This re-scaling of $a$ and shift of $b$, along with the input normalization, allows us to heuristically set the intercept of the linear control signal $b$ to $1$, thereby slightly reducing the computational burden during hyperparameter optimization. 

\section{Predicting the Tipping Point and Post-Tipping-Point Dynamics in Non-Stationary Systems}

To investigate the possible ability of our proposed ML method to predict the dynamics of an unknown non-stationary dynamical system from time series of past system states and generalize by extrapolating to regions of the system state space not explored, or only sparsely explored, in the training data, we consider numerical examples of tipping points associated with the crossing of (1) a saddle-node-induced intermittency bifurcation \cite{pomeau}, (2) an interior crisis, and (3) a subcritical Hopf bifurcation. For the saddle-node bifurcation case, we will use as examples tests on the three-dimensional Lorenz system \cite{lorenz} and the Kuramoto-Sivashinsky partial differential equation \cite{kuramoto,siva}. In particular, we will consider the situation where the non-stationary unknown system initially exhibits periodic motion and at a later time abruptly tips into a chaotic state. For the interior crisis example, we will use as our example the Ikeda map \cite{hammel}, in which the non-stationary unknown system initially evolves on a smaller chaotic attractor that suddenly explodes into a larger chaotic attractor. For the subcritical Hopf bifurcation example, we will consider the Lorenz system, in which the pre-tipping-point orbit moves along a slowly-drifting fixed point attractor of the corresponding stationary system, while the post-tipping-point orbit motion is chaotic and explores a much larger region of the system state space. For all examples, we will consider the case where the non-stationarity of the system is modeled as due to a time-dependent system parameter, and, because we would like to demonstrate our methods for cases more general than a linear drift in the time-dependent system parameter, we will choose the following nonlinear time-dependence for a system parameter $\tilde{\rho}$:

\begin{equation}
    \label{eqn:exp_param}
    \tilde{\rho}(t) = \tilde{\rho}_0 + \tilde{\rho}_1 \exp{t/\tau},
\end{equation}

where $\tilde{\rho}_0$, $\tilde{\rho}_1$, and $\tau$ are constants, and we will consider examples with various settings for these parameters.

For each example, to judge the quality of the ML predictions, we will do the following. We will numerically simulate an ensemble of trajectories of the system of interest from randomly chosen initial conditions in the far past. Next, we will separately train the ML model on each of the trajectories [for $t \in [-t_t,0]$, and, using an appropriate set of hyperparameters (obtained using the optimization scheme described in Sec. II)], and we will generate predictions for $t>0$. We then use the ensemble of true trajectories and ML-predicted trajectories to calculate the Wasserstein-based climate error metric, $\Gamma(t)$, developed in Ref \cite{patel} for some observable quantity $x$ of the system, 

\begin{equation}
    \label{eqn:gamma}
    \Gamma(t) = \frac{2}{\Delta_x} \int|f_a (x,t) - f_p (x,t)|dx,
\end{equation}

where $f_a (x,t)$ is an approximation to the cumulative probability distribution of $x$ over a small interval $[t, t + \delta t]$ obtained from the ensemble of actual trajectories, $f_p (x,t)$ is the cumulative probability distribution of the ensemble of ML-predicted trajectories, and $\Delta_x=x_{max}-x_{min}$ is the range in $x$ from the observed data. See Ref \cite{patel} for a more detailed discussion. 

\subsection{Predicting tipping in the Lorenz System}

We consider the non-stationary, potentially noisy, Lorenz system,

\begin{subequations}
\label{eqn:lorenz}
\begin{equation}
    \frac{dx}{dt} = \sigma (y - x) + \xi_x(t),
\end{equation}
\begin{equation}
    \frac{dy}{dt} = x(\rho(t) - z) - y + \xi_y(t),
\end{equation}
\begin{equation}
    \frac{dz}{dt} = xy - \beta z + \xi_z(t),
\end{equation}
\end{subequations}

where $\sigma = 10$, $\beta = 8/3$, and $\rho$ are the system parameters, and $\xi_i(t)$ for $i=x,y,z$ represents uncorrelated, white (in time) dynamical noise. The dynamical noise is implemented by randomly choosing a number from a uniform distribution over the interval $[-\Bar{\xi}_i,\Bar{\xi}_i]$ and assigning it to $\xi_i$ at each step of the numerical integration of Eqs. (\ref{eqn:lorenz}) (carried out using the fourth order Runge-Kutta method and an integration time step of $\Delta t = 0.01$). The non-stationarity is due to the time-dependent parameter $\rho(t)$ in Eq. (\ref{eqn:lorenz}b) which varies in time according to Eq. (\ref{eqn:exp_param}).

\begin{figure}[h]
    \centering
    \includegraphics[scale=0.5]{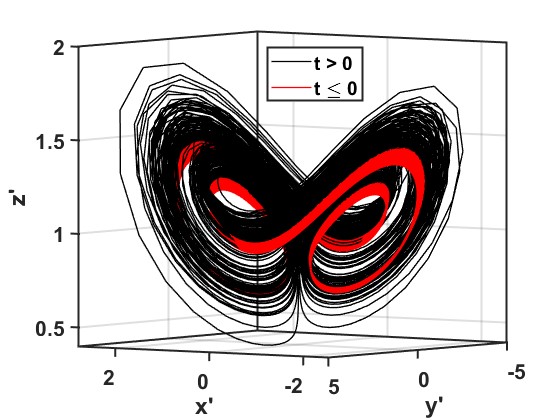}
    \caption{The three-dimensional trajectory of the noiseless non-stationary Lorenz system (given by Eqs. (\ref{eqn:lorenz}) for $\rho_0=154$, $\rho_1=8$, and $\tau=100$ over the time interval $t\in[-600, 120]$) from a randomly chosen initial condition. The red and black curves correspond to the periodic trajectory for $t\leq0$ and the chaotic trajectory for $t>0$, respectively.}
    \label{fig:lorenz_typical_ns}
\end{figure}

First, we consider the noiseless case (i.e., $\xi_i$ = 0 for all $i$), and we set $\rho_0=154$, $\rho_1=8$, and $\tau=100$. Figure \ref{fig:lorenz_typical_ns} shows an example of a non-stationary Lorenz trajectory over $t\in[-600,200]$, starting from a randomly chosen initial condition, for the above choice of parameters. The red and black curves correspond to the trajectory for $t\leq0$ and $t>0$, respectively. For $t<0$, the system motion is characterized by periodic motion of the corresponding stationary system at each time $t<0$, and, at a later time for $t>0$, when the time-dependent parameter $\rho(t)$ crosses a tipping point associated with a saddle-node-induced intermittency bifurcation of the stationary system near $\rho=166$, the motion transitions rapidly to chaotic, which results in the system exploring a larger region of its state space previously unexplored during the periodic motion. In the plot shown in Fig. \ref{fig:lorenz_typical_ns}, we have normalized each variable to the root-mean-square of that variable calculated for $t<0$, and we denote the normalized variables $x', y'$, and $z'$.

\begin{figure}[h]
    \centering
    \includegraphics[scale=0.45]{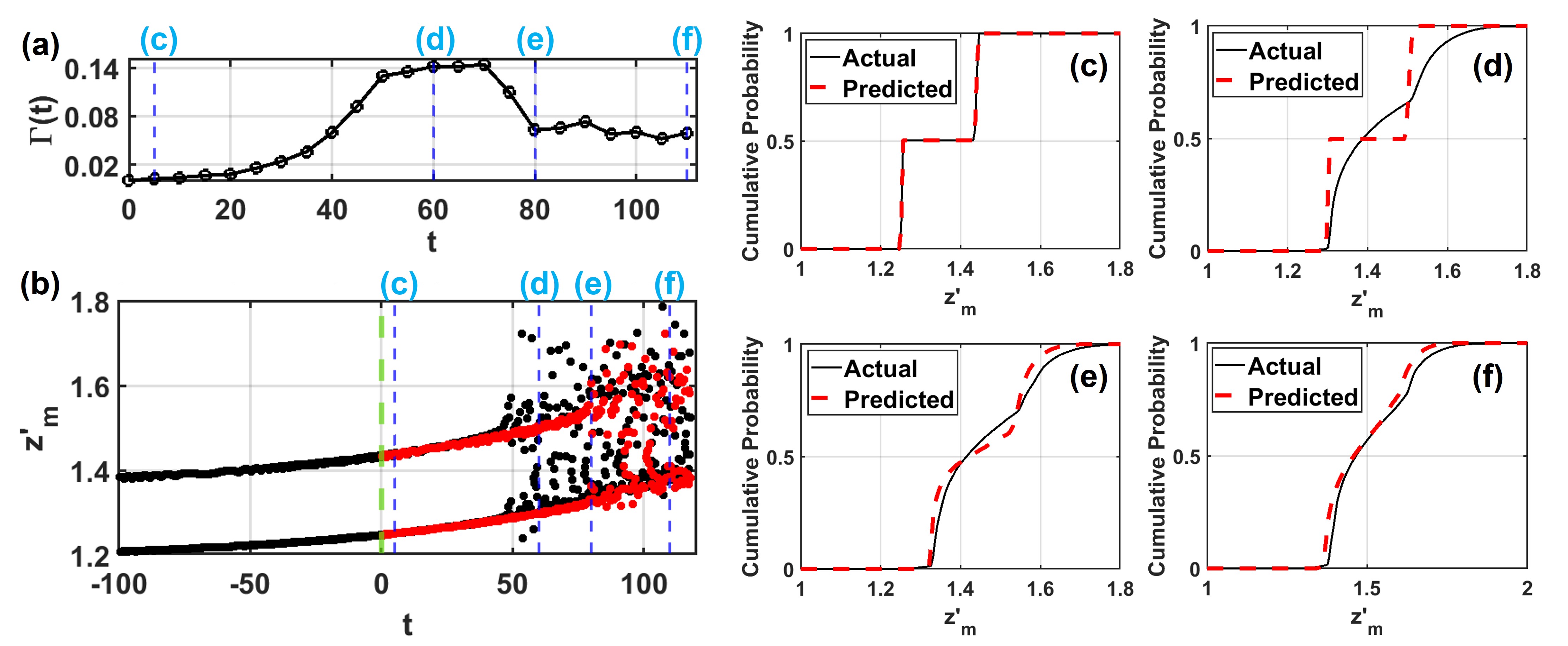}
    \caption{(a) $\Gamma(t)$ versus time over the prediction window. Vertical dashed blue lines labeled (c), (d), (e), and (f) indicate the locations for which the true and predicted $z'_m$ point cumulative probability distributions are shown in panels (c), (d), (e), and (f), respectively. (b) The $z'_m$ points of an example true system trajectory (black dots) and the corresponding ML prediction (red dots). The vertical dashed green line indicates the starting point of the ML prediction. The vertical dashed blue lines indicate the same times as in panel (a). Panels (c), (d), (e), and (f) show the true (solid black curves) and ML-predicted (dashed red curves) $z'_m$ cumulative probability distributions at $t=5,60,80$, and $110$, respectively.}
    \label{fig:lorenz_noiseless_tau100}
\end{figure}

Figure \ref{fig:lorenz_noiseless_tau100} shows results of the application of our ML technique to the above non-stationary noiseless Lorenz case. Figure \ref{fig:lorenz_noiseless_tau100}(a) shows the climate error metric $\Gamma(t)$ (see Eq. (\ref{eqn:gamma})), obtained using an ensemble of $1000$ trajectories from randomly chosen initial conditions, over the prediction window. The observables we use to construct this metric are the maxima of $z'(t)$, denoted $z'_m$ where the subscript $m$ labels the $m^{th}$ maximum. Figure \ref{fig:lorenz_noiseless_tau100}(b) shows a typical example of the points $z'_m$ versus $t$ from a numerically integrated (``true") trajectory (black dots), and the ML prediction for $t>0$ (red dots) after training on the true system trajectory segment (for $t<0$) shown in the plot. In this case, by a ``typical" trajectory, we mean a trajectory from a randomly chosen initial condition. The vertical dashed green line in panel (b) indicates the starting point of the ML prediction. We see that, similar to the true trajectory, the ML-predicted trajectory initially evolves as a periodic orbit and then at a later time rapidly transitions to a chaotic orbit which explores a large state space region. The vertical dashed blue lines in Figs. \ref{fig:lorenz_noiseless_tau100}(a) and \ref{fig:lorenz_noiseless_tau100}(b) at $t=5, 60, 80$, and $110$, labeled by (c), (d), (e), and (f), respectively, denote the times for which we show the true ($f_a(z'_m,t)$) and predicted ($f_p(z'_m,t)$) cumulative distributions of the $z'_m$ points. These distributions are shown in panels (c), (d), (e), and (f) , respectively. In Fig. \ref{fig:lorenz_noiseless_tau100}(c) the two-step structure of the predicted (dashed red curve) cumulative probability distribution of the $z'_m$ points indicates that the predicted ensemble of trajectories are periodic (with period $2$) near $t=5$, and that the predicted distribution is in agreement with the true distribution (solid black curve). Figure \ref{fig:lorenz_noiseless_tau100}(d) shows the cumulative probability distributions at $t=60$. It is seen that, at this time, whereas the true system trajectories have undergone a tipping point from periodic to chaotic motion, the ensemble of predicted trajectories are still periodic. This discrepancy is also what gives rise to the peak in $\Gamma(t)$ seen in panel (a). Figures \ref{fig:lorenz_noiseless_tau100}(e) and \ref{fig:lorenz_noiseless_tau100}(f) show the true and predicted cumulative probability distributions at later times ($t=80$, and $110$, respectively), from which we see that the ensemble of predicted trajectories have also passed a tipping point and transitioned from periodic to chaotic motion. Accordingly, the value of $\Gamma(t)$ decreases from the spike near $t=60$. Considering the small values of the climate error metric in Fig. \ref{fig:lorenz_noiseless_tau100}(a) (except for the spike due to the error in the ML prediction of the exact timing of the tipping point transition), and comparing the pre- and post-transition cumulative distributions in Figs. \ref{fig:lorenz_noiseless_tau100}(c) and \ref{fig:lorenz_noiseless_tau100}(e), we judge that the ML-predicted trajectories not only anticipate a tipping point (although with some error in the tipping point time), but also accurately predict the post-tipping-point dynamics. We regard this as impressive since the ML model was only trained on time series of the true system trajectory for $t<0$, which corresponds to a periodic motion that explores a restricted region of state space, but still accurately predicts the post-tipping-point chaotic dynamics. Thus the ML model is able to extrapolate to a previously unexplored region of the system state space. Table \ref{tab:Lorenz_noiseless_tau100} shows the hyperparameters for this example, obtained using the hyperparameter optimization scheme described in Sec. 2.2.

\begin{table}[h]
    \centering
    \caption{Hyperparameters for predicting the noiseless non-stationary Lorenz system in Fig. \ref{fig:lorenz_noiseless_tau100}.}
    \begin{tabular}{m{25em}|m{4em}|m{6em}}
         Reservoir size & $N$ & $2000$  \\
         \hline
         Average number of connections per node & $\langle d \rangle$ & $3$\\
         \hline
         Spectral radius of the reservoir's adjacency matrix & $\rho_s$ & $0.8$\\
         \hline
         Coupling strength of the input to the reservoir & $\chi$ & $0.5$\\
         \hline
         Reservoir leakage parameter & $\alpha$ & $1$\\
         \hline
         Reservoir node activation bias & $b_r$ & $0$\\
         \hline
         Tikhonov regularization parameter & $\alpha$ & $3.59\times 10^{-13}$\\
         \hline
         Intercept of the linear control signal & $b$ & $1$\\
         \hline
         Slope of the linear control signal & $a$ & $5.99\times 10^{-5}$\\
         \hline
         Numerical integration time step for Eq. (\ref{eqn:lorenz}) & $\Delta t$ & $0.01$\\
         \hline
         RC time step & $\Delta t_{RC}$ & $0.01$\\
         \hline
         Strength of observational noise added to training data & $\epsilon_0$ & $5\times 10^{-5}$\\
         \hline
         Number of passes of training data during training & &$10$\\
    \end{tabular}
    \label{tab:Lorenz_noiseless_tau100}
\end{table}

We now discuss conditions that potentially enable the ML to generalize and extrapolate its learning to unexplored regions of the target system state space in the above example, as well as situations in which we may not expect such predictive capabilities, and some possible elements which may alleviate the latter case to some extent. In the example (Fig. \ref{fig:lorenz_noiseless_tau100}) the orbit is exactly periodic in a corresponding stationary situation. However, due to the non-stationarity, the location of the periodic orbit drifts with time, giving the non-stationary orbit a ``thickness" as seen by the red curve in Fig. \ref{fig:lorenz_typical_ns}. This allows the trajectory to sample the system state space in a ``band" region rather than only on the closed curve periodic orbit that would apply if the underlying system was stationary. As seen from the example in Fig. \ref{fig:lorenz_noiseless_tau100}, this evidently allows the ML to learn the changing dynamics and to extrapolate from it so as to predict both a tipping point as well as the post-tipping-point dynamics which explore a larger region of the system state space. Based on this consideration, a natural question is what would happen if the parameter-sweep (i.e., the change in the non-stationary-inducing time-dependent system parameter) over the same training time duration were significantly smaller (e.g., if the time-scale of non-stationarity were significantly larger than the time interval over which the system was observed for creating the training dataset)? As shown below, we find that, in the case where the observed periodic orbit (in the training data) does not have sufficient parameter-sweep (and the system is evolving in the absence of dynamical noise), the ML is unable to anticipate a tipping-point transition from periodic to chaotic motion. However, if the same system (over the same insufficient parameter-sweep) is evolving in the presence of dynamical noise of sufficient strength, then the ML is able to reliably anticipate a tipping point transition, and in many cases, predict the post-tipping-point transition dynamics.

\begin{figure}[h]
    \centering
    \includegraphics[scale=0.45]{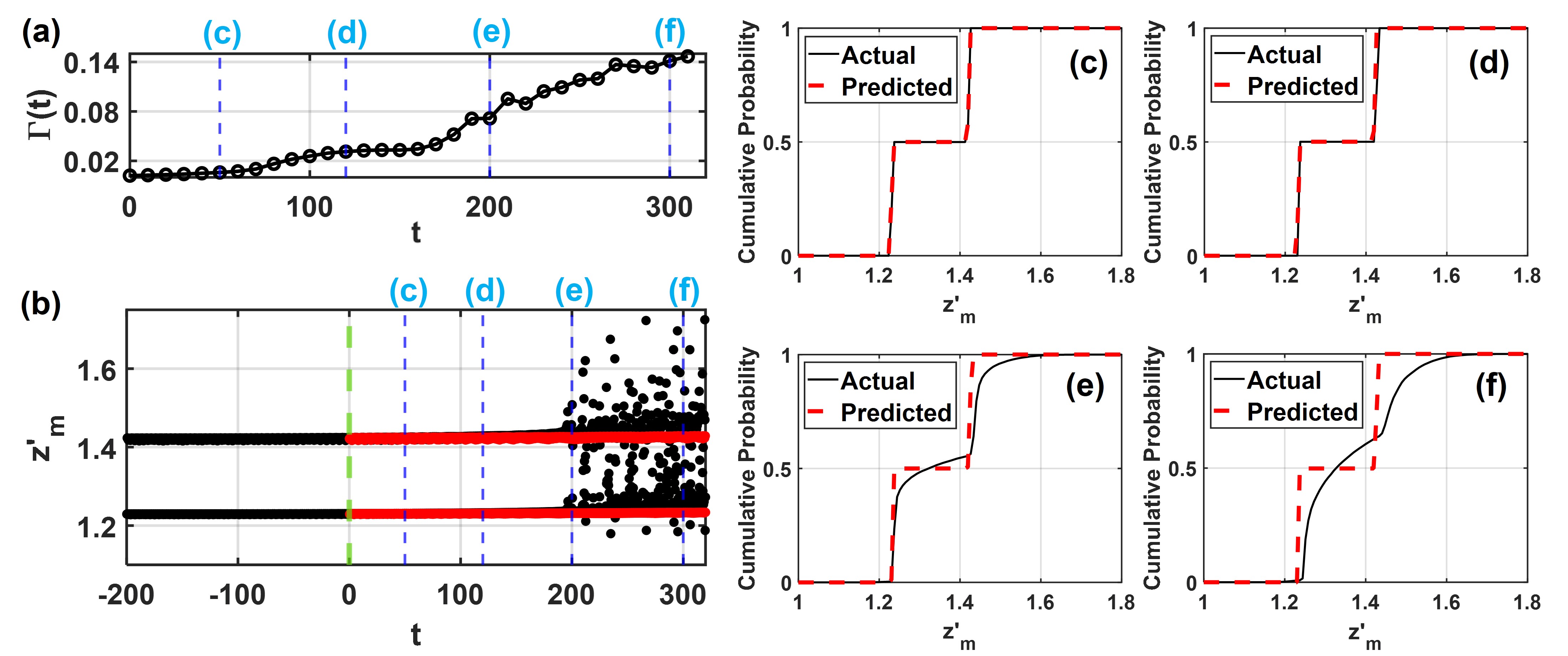}
    \caption{(a) $\Gamma(t)$ versus time over the prediction window. Vertical dashed blue lines labeled (c), (d), (e), and (f) indicate the locations for which the true and predicted $z'_m$ point cumulative probability distributions are shown in panels (c), (d), (e), and (f), respectively. (b) The $z'_m$ points of an example true system trajectory (black dots) and the corresponding ML prediction (red dots). The vertical dashed green line indicates the starting point of the ML prediction. The vertical dashed blue lines indicate the same as in panel (a). Panels (c), (d), (e), and (f) show the true (solid black curves) and ML-predicted (dashed red curves) $z'_m$ point cumulative probability distributions at $t=50,120,200$, and $300$, respectively.}
    \label{fig:lorenz_isps_noiseless_tau100}
\end{figure}

We now illustrate the above points. First we consider the noiseless non-stationary Lorenz system given by Eq. (\ref{eqn:lorenz}), and $\rho_0=165.5$, $\rho_1=0.1$, and $\tau=100$. In contrast, for the case illustrated in Figs. \ref{fig:lorenz_typical_ns} and \ref{fig:lorenz_noiseless_tau100}, $\rho_0 = 154$, $\rho_1 = 8$, and the total range of the parameter drift over the same training time duration is much reduced. As with the previous example, for $t<0$ the system orbit is periodic and at a later time (for $t>0$) the parameter $\rho$ crosses a tipping point near $\rho=166$, after which the system orbits become chaotic. Figure \ref{fig:lorenz_isps_noiseless_tau100} shows the results of applying our ML methods to an ensemble of $1000$ trajectories from randomly chosen initial conditions. Figure \ref{fig:lorenz_isps_noiseless_tau100}(a) shows $\Gamma(t)$ over the prediction window and the vertical dashed blue lines indicate the times for which the true and ML-predicted $z'_m$ point cumulative probability distributions have been plotted in the figure panels corresponding to the labels. Figure \ref{fig:lorenz_isps_noiseless_tau100}(b)  shows the $z'_m$ points of an example true system trajectory (black dots) and of the corresponding ML-predicted trajectory (red dots). We see that, whereas the true system trajectory undergoes a tipping-point transition from periodic to chaotic motion near $t=200$, the ML-predicted trajectory does not pass through a transition and instead continues to show periodic motion. The cumulative probability distributions shown in Figs. \ref{fig:lorenz_isps_noiseless_tau100}(c), \ref{fig:lorenz_isps_noiseless_tau100}(d), \ref{fig:lorenz_isps_noiseless_tau100}(e), \ref{fig:lorenz_isps_noiseless_tau100}(f) at $t=50,120,200$, and $300$, respectively, indicate that the ensemble of ML-predicted trajectories fail to undergo a tipping point from periodic to chaotic motion. We attribute this to the parameter-sweep of the system over the training data being too small to allow extrapolation by our ML method into the previously unexplored state space regions. Table \ref{tab:Lorenz_isps_noiseless_tau100} shows the hyperparameters used for this example.

\begin{table}[h]
    \centering
    \caption{Hyperparameters for predicting the noiseless non-stationary Lorenz system (in Fig. \ref{fig:lorenz_isps_noiseless_tau100}) with an insufficient parameter-sweep.}
    \begin{tabular}{m{25em}|m{4em}|m{6em}}
         Reservoir size & $N$ & $2000$  \\
         \hline
         Average number of connections per node & $\langle d \rangle$ & $3$\\
         \hline
         Spectral radius of the reservoir's adjacency matrix & $\rho_s$ & $1$\\
         \hline
         Coupling strength of the input to the reservoir & $\chi$ & $1$\\
         \hline
         Reservoir leakage parameter & $\alpha$ & $0.5$\\
         \hline
         Reservoir node activation bias & $b_r$ & $0$\\
         \hline
         Tikhonov regularization parameter & $\alpha$ & $1.29\times 10^{-12}$\\
         \hline
         Intercept of the linear control signal & $b$ & $1$\\
         \hline
         Slope of the linear control signal & $a$ & $1\times 10^{-6}$\\
         \hline
         Numerical integration time step for Eq. (\ref{eqn:lorenz}) & $\Delta t$ & $0.01$\\
         \hline
         RC time step & $\Delta t_{RC}$ & $0.01$\\
         \hline
         Strength of observational noise added to training data & $\epsilon_0$ & $5\times 10^{-4}$\\
         \hline
         Number of passes of training data during training & &$10$\\
    \end{tabular}
    \label{tab:Lorenz_isps_noiseless_tau100}
\end{table}

\begin{figure}[h]
    \centering
    \includegraphics[scale=0.45]{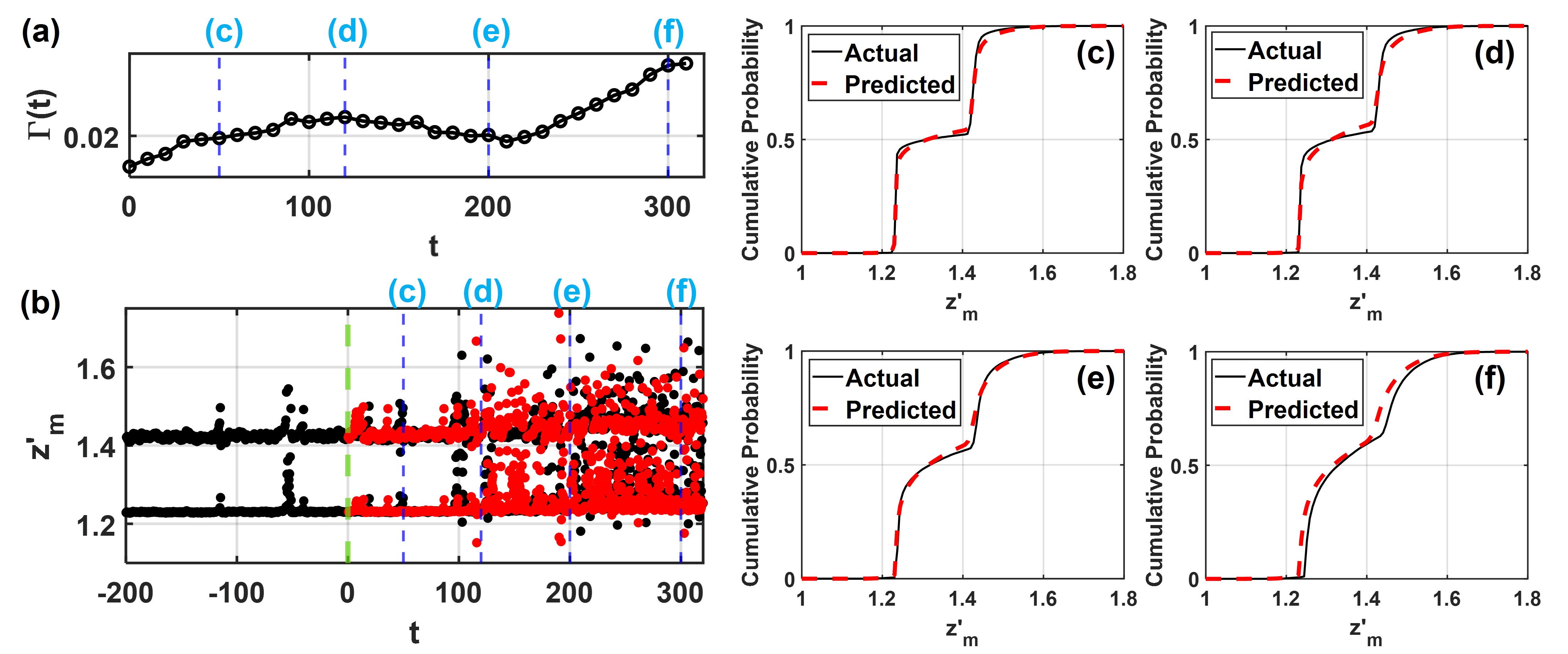}
    \caption{(a) $\Gamma(t)$ vs time over the prediction window. Vertical dashed blue lines labeled (c), (d), (e), and (f) indicate the locations for which the true and predicted $z'_m$ point cumulative probability distributions are shown in panels (c), (d), (e), and (f), respectively. (b) The $z'_m$ points of an example true system trajectory (black dots) and the corresponding ML prediction (red dots). The vertical dashed green line indicates the starting point of the ML prediction. The vertical dashed blue lines indicate the same as in panel (a). Panels (c), (d), (e), and (f) show the true (solid black curves) and ML-predicted (dashed red curves) $z'_m$ map point cumulative probability distributions at $t=50,120,200$, and $300$, respectively.}
    \label{fig:lorenz_isps_noisy_tau100}
\end{figure}

Now we consider the above example, but where the system is evolving in the presence of dynamical noise of strength $\bar{\xi}_i = 3\times 10^{-2}$ for $i=x,y,z$. Figure \ref{fig:lorenz_isps_noisy_tau100} shows the results of applying our ML methods to this example using an ensemble of $1000$ trajectories from randomly chosen initial conditions. Figure \ref{fig:lorenz_isps_noisy_tau100}(a) shows $\Gamma(t)$ over the prediction window, and Fig. \ref{fig:lorenz_isps_noisy_tau100}(b) shows the $z'_m$ points from a typical true system trajectory (black dots) and the corresponding ML-predicted trajectory (red dots) from an ML model trained on the true trajectory shown (for $t<0$). We see that, unlike in the noiseless case, the ML-predicted trajectory undergoes a tipping-point transition from (noisy) periodic to chaotic motion. In addition, we see that the  ML-predicted trajectory captures the intermittency bursting behavior before the tipping point due to the dynamical noise. Figures \ref{fig:lorenz_isps_noisy_tau100}(c), \ref{fig:lorenz_isps_noisy_tau100}(d), \ref{fig:lorenz_isps_noisy_tau100}(e), \ref{fig:lorenz_isps_noisy_tau100}(f) show the true and ML-predicted cumulative probability distributions of the $z'_m$ points at $t=50,120,200$, and $t=300$, respectively. From these distributions, it is seen that the ML-predicted ensemble of trajectories initially exhibit a noisy periodic motion and then transition to a different motion which closely approximates that of the true system trajectories. Through this example with small parameter sweep, we see that although the ML was unable to anticipate a tipping-point transition from periodic to chaotic motion in the case of the noiseless Lorenz system, the presence of dynamical noise of sufficient strength in that same system enabled the ML (trained using the same length of data as in the noiseless case) to anticipate a tipping-point transition and predict the post-transition system dynamics. (Table \ref{tab:Lorenz_isps_noisy_tau100} shows the hyperparameters used for this example.) The interpretation is that, by kicking the orbit off its period $2$ trajectory during the training, the dynamical noise allows the ML to sample the system dynamics in the neighborhood of the periodic orbit, and the ML is then able to use this added information to accomplish its prediction task.

\begin{table}[h]
    \centering
    \caption{Hyperparameters for predicting the noisy non-stationary Lorenz system (in Fig. \ref{fig:lorenz_isps_noisy_tau100}) with an insufficient parameter-sweep.}
    \begin{tabular}{m{25em}|m{4em}|m{6em}}
         Reservoir size & $N$ & $2000$  \\
         \hline
         Average number of connections per node & $\langle d \rangle$ & $3$\\
         \hline
         Spectral radius of the reservoir's adjacency matrix & $\rho_s$ & $0.75$\\
         \hline
         Coupling strength of the input to the reservoir & $\chi$ & $0.5$\\
         \hline
         Reservoir leakage parameter & $\alpha$ & $0.5$\\
         \hline
         Reservoir node activation bias & $b_r$ & $0$\\
         \hline
         Tikhonov regularization parameter & $\alpha$ & $1\times 10^{-13}$\\
         \hline
         Intercept of the linear control signal & $b$ & $1$\\
         \hline
         Slope of the linear control signal & $a$ & $2.78\times 10^{-6}$\\
         \hline
         Numerical integration time step for Eq. (\ref{eqn:lorenz}) & $\Delta t$ & $0.01$\\
         \hline
         RC time step & $\Delta t_{RC}$ & $0.01$\\
         \hline
         Strength of observational noise added to training data & $\epsilon_0$ & $0$\\
         \hline
         Number of passes of training data during training & &$1$\\
    \end{tabular}
    \label{tab:Lorenz_isps_noisy_tau100}
\end{table}

\subsection{Predicting tipping points and evolution of extreme event frequency in the non-stationary Ikeda Map}

\begin{figure}[h]
    \centering
    \includegraphics[scale=0.55]{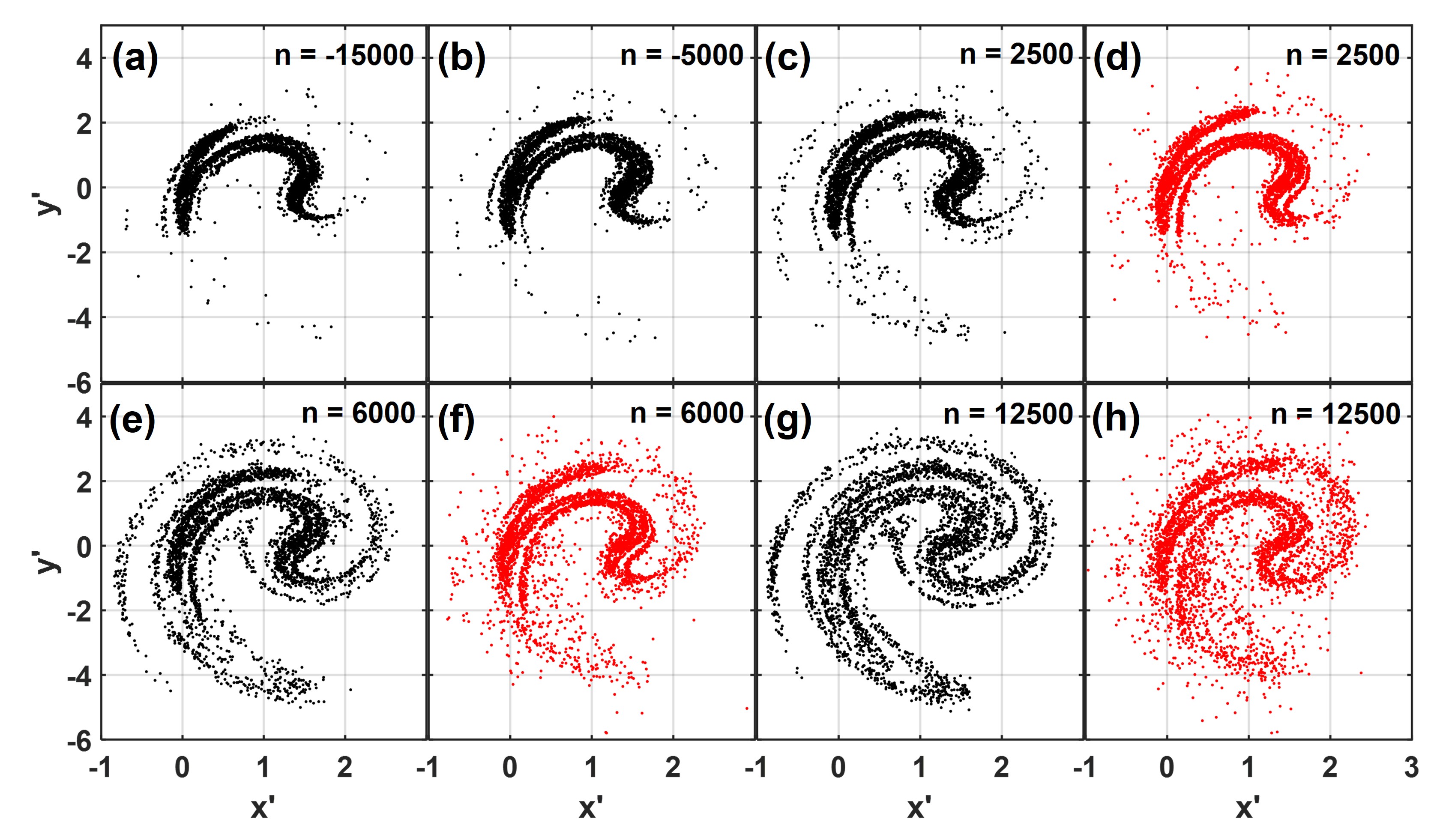}
    \caption{Phase portraits of the snapshot attractor of the true system at (a) $n=-15000$, (b) $n=-5000$, (c) $n=2500$, (e) $n=6000$, (g) $n=12500$ and corresponding ML prediction at (d) $n=2500$, (f) $n=6000$, and (h) $n=125000$.}
    \label{fig:ikeda_ssa}
\end{figure}

\begin{figure}[h]
    \centering
    \includegraphics[scale=0.45]{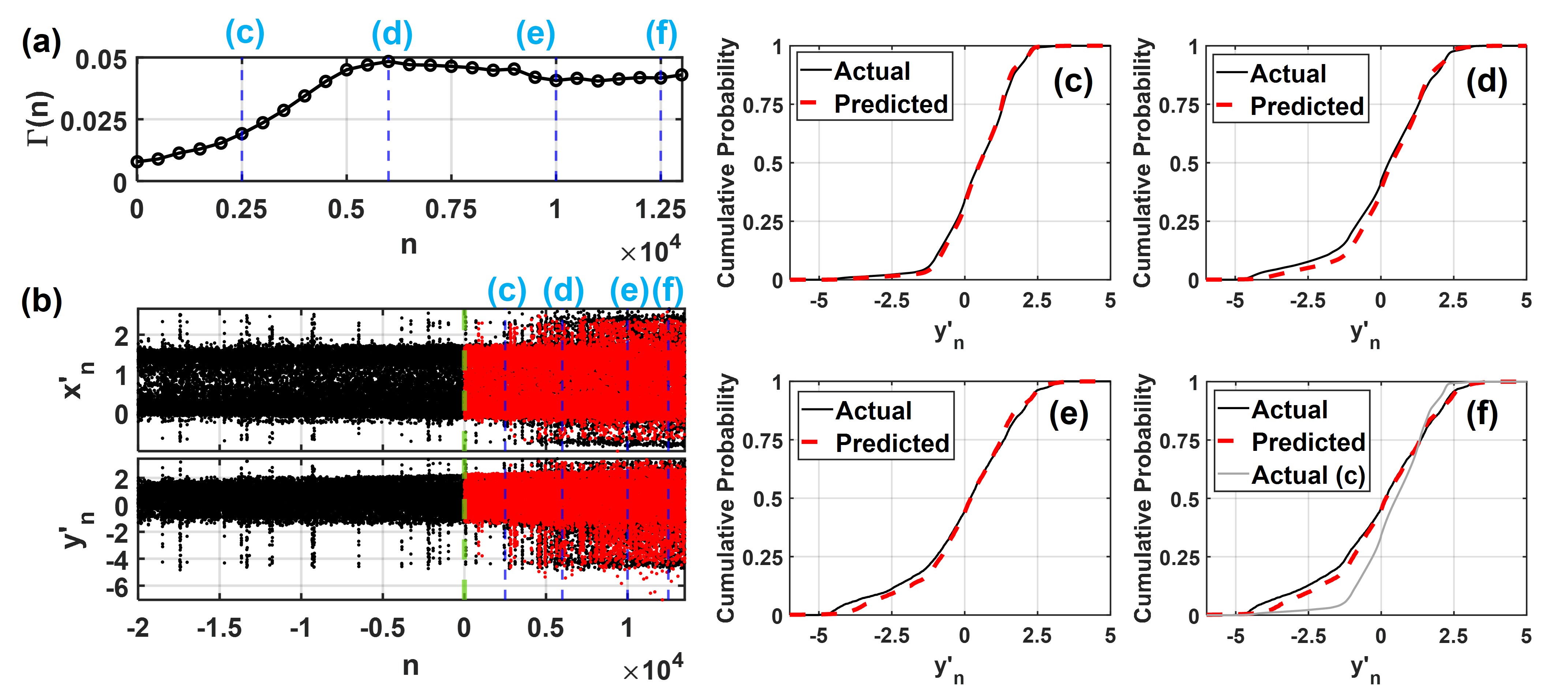}
    \caption{(a) $\Gamma(n)$ vs $n$ over the prediction window. Vertical dashed blue lines labeled (c), (d), (e), and (f) indicate the locations for which the true and predicted $y'_n$ cumulative probability distributions are shown in panels (c), (d), (e), and (f), respectively. (b) $x'_n$ (top) and $y'_n$ (bottom) of an example true system trajectory (black dots) and the corresponding ML prediction (red dots). The vertical dashed green line indicates the starting point of the ML prediction. The vertical dashed blue lines indicate the same as in panel (a). Panels (c), (d), (e), and (f) show the true (solid black curves) and ML-predicted (dashed red curves) $y'_n$ cumulative probability distributions at $n=2500,6000,10000$, and $12500$, respectively.}
    \label{fig:ikeda_gamma_cdf}
\end{figure}

Interior crisis transitions have been identified as a common route to rare extreme events in a variety of dynamical systems \cite{mishra}. In non-stationary systems that have crossed a tipping point mediated by an interior crisis transition of the corresponding stationary system, the frequency of rare extreme events increases as the bifurcation parameter drifts further past the critical value. As this process continues, the system eventually enters a regime where the previously rare extreme events become part of the regular dynamics leading to much greater persistent variability of the system state (e.g., see Fig. \ref{fig:ikeda_ssa}). In noisy systems the rare and extreme events can occur before the tipping point is crossed (e.g., see Figs. \ref{fig:ikeda_ssa}(a,b,c) and \ref{fig:ikeda_gamma_cdf}(b)). The task of predicting such tipping point and post-tipping-point behavior has become more urgent in recent years, e.g., due to concerns of potential increase in the rate of extreme weather such as heavy rainfall \cite{trenberth,goswami}, floods, and hurricanes driven by a warming terrestrial climate. 

Next, we consider prediction of the interior crisis, and the climatic variation of large amplitude events induced by an interior crisis and dynamical noise in the non-stationary noisy Ikeda map \cite{ray} (represented here as a $2$-dimensional map of real variables $x_n$ and $y_n$),

\begin{subequations}
\label{eqn:ikeda}
\begin{equation}
    x_{n+1} = a + b(x_n \cos(\gamma - \frac{\eta_n}{c_n}) - y_n \sin(\gamma - \frac{\eta_n}{c_n})) + \xi_{x,n},
\end{equation}
\begin{equation}
    y_{n+1} = b(x_n \sin(\gamma - \frac{\eta_n}{c_n}) + y_n \cos(\gamma - \frac{\eta_n}{c_n})) + \xi_{y,n},
\end{equation}
\end{subequations}

where $c_n=1+x_n^2+y_n^2$, $\gamma$, $a$, and $b$ are real constants, $\eta_n$ is a real time-dependent parameter, and $\xi_{x,n}$ and $\xi_{y,n}$ are independent temporally-uncorrelated dynamical noise terms. 

We choose $a=0.85$, $b=0.9$, and $\gamma=0.4$. First, we briefly consider the noiseless stationary system behavior near the interior crisis which occurs at $\eta = \eta_c \approx 7.269$. For $\eta$ slightly less than $\eta_c$, typical system orbits evolve on a smaller chaotic attractor. As $\eta$ is increased above $\eta_c$, typical system orbits start to exhibit signatures of extreme events, i.e., after spending some time evolving on the smaller chaotic attractor they suddenly move far from it before returning. The frequency with which such bursts occur increases as $\eta$ increases until orbits on the large post-crisis chaotic attractor densely cover the expanded attractor region. 

In our noisy non-stationary example, we choose the same parameter values for $a$, $b$, and $\gamma$, and the non-stationarity is induced by the following time-dependence of $\eta$,

\begin{equation}
    \eta_n = \eta^{(0)} + \eta^{(1)} \exp{(n/\tau)}
\end{equation}

where $\eta^{(0)}=6.75$, $\eta^{(1)}=0.5$, and $\tau=20000$. The dynamical noise is implemented by randomly choosing a number from a uniform distribution over $[-0.015, 0.015]$ and assigning it to $\xi_i$ for $i=x,y$ at each iteration of Eqs. (\ref{eqn:ikeda}). For this choice of system parameters, before the start of prediction ($n<0$) typical system orbits are mostly confined to the region of system state space corresponding to the smaller pre-crisis chaotic attractor of the stationary system, with occasional bursting due to the presence of dynamical noise. At later times $n>0$, after the system has undergone a tipping point transition due to the interior crisis, typical system orbits much more regularly explore the larger region of system state space corresponding to the larger post-crisis chaotic attractor of the stationary system. Similar to the Lorenz system example, we assume we have observed the system state for $n<0$, when the system motion is largely confined to a smaller region of the state space. We note that although there is noise-induced bursting in typical system orbits for $n<0$ (i.e., in the training data), for the large majority of time the orbit is in the smaller pre-crisis region. This can be seen in Figs. \ref{fig:ikeda_ssa}(a) and \ref{fig:ikeda_ssa}(b), which show the snapshot attractors of the system at $n=-15000$ and $n=-5000$, respectively. The snapshot attractors are constructed by plotting the system state at a given time $n$ of an ensemble of $4000$ trajectories initialized in the far past from randomly chosen initial conditions. Such plots give a meaningful measure of the distribution of states of the non-stationary, noisy system at a given time $n$. From Figs. \ref{fig:ikeda_ssa}(a) and \ref{fig:ikeda_ssa}(b) we see that most of the points are contained within the region of state space corresponding to the pre-crisis, noiseless, chaotic attractor of the stationary system. This means that any information the training data contains regarding the larger region of state space explored by the post-tipping-point system orbits is purely due to the presence of dynamical noise and, when this noise is small such information can be very limited, but it is nevertheless crucial as we will discuss shortly. Here, $x'_n$ and $y'_n$ are obtained by normalizing $x_n$ and $y_n$, respectively, by the root-mean-square of their values for $n<0$.

Figure \ref{fig:ikeda_gamma_cdf}(a) shows $\Gamma(n)$ over the prediction window, calculated from the ensemble of $4000$ trajectories, and Fig. \ref{fig:ikeda_gamma_cdf}(b) shows an example of a surface of section plots for a typical system orbit (black dots) and the corresponding ML-predicted orbit for $n>0$ (red dots). We see from Fig. \ref{fig:ikeda_gamma_cdf}(b) that the ML-predicted orbit, similar to the true orbit, initially evolves mostly on the smaller region of state space with occasional bursting, and then near $n=5000$ undergoes a tipping point transition, after  which the orbit begins to evolve on the larger post-tipping-point region of state space more frequently. This is further supported by the snapshot attractors at $n=2500$ and $n=6000$ in Figs.\ref{fig:ikeda_ssa}(c), \ref{fig:ikeda_ssa}(d), \ref{fig:ikeda_ssa}(e), and \ref{fig:ikeda_ssa}(f) generated by the true and ML-predicted ensemble of trajectories which show the true and predicted pre-tipping-point and post-tipping-point behavior. We see that at $n=2500$ the true and ML-predicted ensemble of trajectories spend significantly less time in the region of state space into which the attractor expands due to the crisis than they do at $n=6000$. The snapshot attractors at $n=12500$, shown in Figs. \ref{fig:ikeda_ssa}(g) and \ref{fig:ikeda_ssa}(h), show that, like the true system orbits, the ensemble of ML-predicted trajectories have undergone a fundamental change in behavior compared to that observed in the training data. This is also captured in the cumulative probability distributions of $y'_n$ in Figs. \ref{fig:ikeda_gamma_cdf}(c), \ref{fig:ikeda_gamma_cdf}(d), \ref{fig:ikeda_gamma_cdf}(e) and \ref{fig:ikeda_gamma_cdf}(f) calculated at $n=2500$, $n=6000$, $n=10000$, and $n=12500$, respectively. Figure \ref{fig:ikeda_gamma_cdf}(c) indicates that the pre-tipping-point ensemble of true and ML-predicted trajectories are mostly restricted to the smaller region of state space $y_n \in [-2.5, 2.5]$ with a thin tail extending beyond $y'_n=-2.5$ due to the occasional extreme events. The cumulative probability distributions in Fig. \ref{fig:ikeda_gamma_cdf}(f) show that the ensemble of true and ML-predicted trajectories explore a larger region of state space than during the pre-tipping-point regime. The cumulative distribution of $y'_n$ of the true ensemble of trajectories at $n=2500$, the case shown in Fig. \ref{fig:ikeda_gamma_cdf}(c), is plotted in Fig \ref{fig:ikeda_gamma_cdf}(f) (solid gray curve) for comparison. We note that, as mentioned above, in this example the training data contains extreme events which result in the system orbit sparsely sampling the larger region of state space explored by the post-tipping-point trajectory. We find, however, that although this sampling is sparse, the limited information obtained about the larger region of state space explored by the post-tipping-point orbits is crucial. We found that the ML was unable to consistently and accurately predict the post-tipping-point transition dynamics of the system when the strength of the dynamical noise was insufficient to cause extreme events in the training data. The presence of extreme events in the training data (due to the presence of dynamical noise of sufficient strength), even if infrequent, drastically improves the ML's ability to predict the post-tipping-point transition system dynamics. (We also note that in this example, unlike in the Lorenz system example, a small fraction of the ML-predicted trajectories in the ensemble (less than $5\%$) became unstable in the post-tipping-point regime (meaning that these predictions diverged to very large values), and these unstable trajectories were not used in calculating the statistics presented in Figs. \ref{fig:ikeda_ssa} and \ref{fig:ikeda_gamma_cdf}.)

To enable accurate climate prediction of the Ikeda map, we used the modified ML architecture described in Appendix B. Table \ref{tab:ikeda} shows the hyperparameters used for the example in this section.

\begin{table}[h]
    \centering
    \caption{Hyperparameters for predicting the Ikeda map.}
    \begin{tabular}{m{25em}|m{4em}|m{6em}}
         Reservoir size & $N$ & $1000$  \\
         \hline
         Average number of connections per node & $\langle d \rangle$ & $3$\\
         \hline
         Spectral radius of the reservoir's adjacency matrix & $\rho_s$ & $0.25$\\
         \hline
         Maximal coupling strength of the dynamical variable input to the reservoir & $\chi_1$ & $2$\\
         \hline
         Maximal coupling strength of the parameter input to the reservoir & $\chi_2$ & $1$\\
         \hline
         Reservoir leakage parameter & $\alpha$ & $1$\\
         \hline
         Reservoir node activation bias & $b$ & $0$\\
         \hline
         Tikhonov regularization parameter & $\alpha$ & $7.74\times 10^{-8}$\\
         \hline
         Intercept of the linear control signal & $b$ & $1$\\
         \hline
         Slope of the linear control signal & $a$ & $4.64\times 10^{-7}$\\
         \hline
         RC time step & $\Delta n$ & $1$\\
         \hline
         Strength of observational noise added to training data & $\epsilon_0$ & $5 \times 10^{-2}$\\
         \hline
         Number of passes of training data during training & &$10$\\
    \end{tabular}
    \label{tab:ikeda}
\end{table}

\subsection{Predicting tipping in the Kuramoto-Sivashinsky equations}

In this section, we use machine learning to anticipate the tipping point transition and predict the post-tipping-point dynamics of an unknown system taken to be the non-stationary, potentially noisy, Kuramoto-Sivashinsky equation,

\begin{equation}
    \label{eqn:ks}
    \frac{\partial w(x,t)}{\partial t} + w(x,t) \frac{\partial w(x,t)}{\partial x} + \frac{\partial^2 w(x,t)}{\partial x^2} + \kappa(t) \frac{\partial^4 w(x,t)}{\partial x^4} = \xi (x,t),
\end{equation}

where $w(x,t)$ is a real-valued scalar function of $x$ and $t$, with periodic boundary conditions $w(x,t) = w(x+2\pi,t)$, $\kappa(t)$ is a time-dependent system parameter, and $\xi(x,t)$ is an uncorrelated and white (in time) dynamical noise term. 

We solve Eq. (\ref{eqn:ks}) numerically using the exponential time-differencing Runge-Kutta fourth-order time-stepping scheme \cite{kassam} with a step size of $\Delta t = 0.0084$ and a spatial grid of $64$ grid points (i.e., $x$ is discretized into $x_i$ for $i=1,2,..,64$). 

\begin{figure}[h]
    \centering
    \includegraphics[scale=0.75]{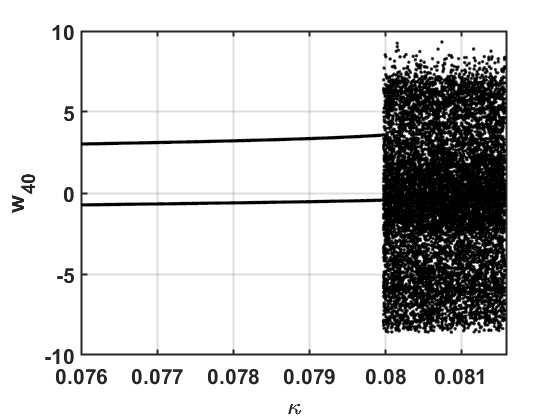}
    \caption{Bifurcation diagram of the \emph{stationary} Kuramoto-Sivashinsky equation for $\kappa \in [0.076, 0.0816]$ obtained by the $w_{40}$-Poincare map points.}
    \label{fig:ks_bifur}
\end{figure}

For the examples below, we consider the periodic window which exists near $\kappa = 0.08$. Figure \ref{fig:ks_bifur} shows a bifurcation diagram of the \emph{stationary} system for $\kappa \in [0.076, 0.0816]$, constructed by using the Poincare surface of section $w(x_1, t) = 0$ with $\partial w(x_1,t) / \partial t > 0$ and observing the value of $w(x_{40},t)$ at many surface of section crossings ($t = t_n$), for different values of $\kappa$. We denote the Poincare surface of section map points by $w_{m}(n) = w(x_{m},t_n)$. We see that for $\kappa \gtrapprox 0.08$, the system dynamics is chaotic. Indeed, as shown in Ref. \cite{edson} and verified by our own numerical calculations (not shown) the largest Lyapunov exponent for this system at $\kappa=0.0816$ is about $0.043$ and the Kaplan-Yorke dimension of the attractor is about $3.2$. For $\kappa \lessapprox 0.08$, the motion is periodic. 

\begin{figure}[h]
    \centering
    \includegraphics[scale=0.45]{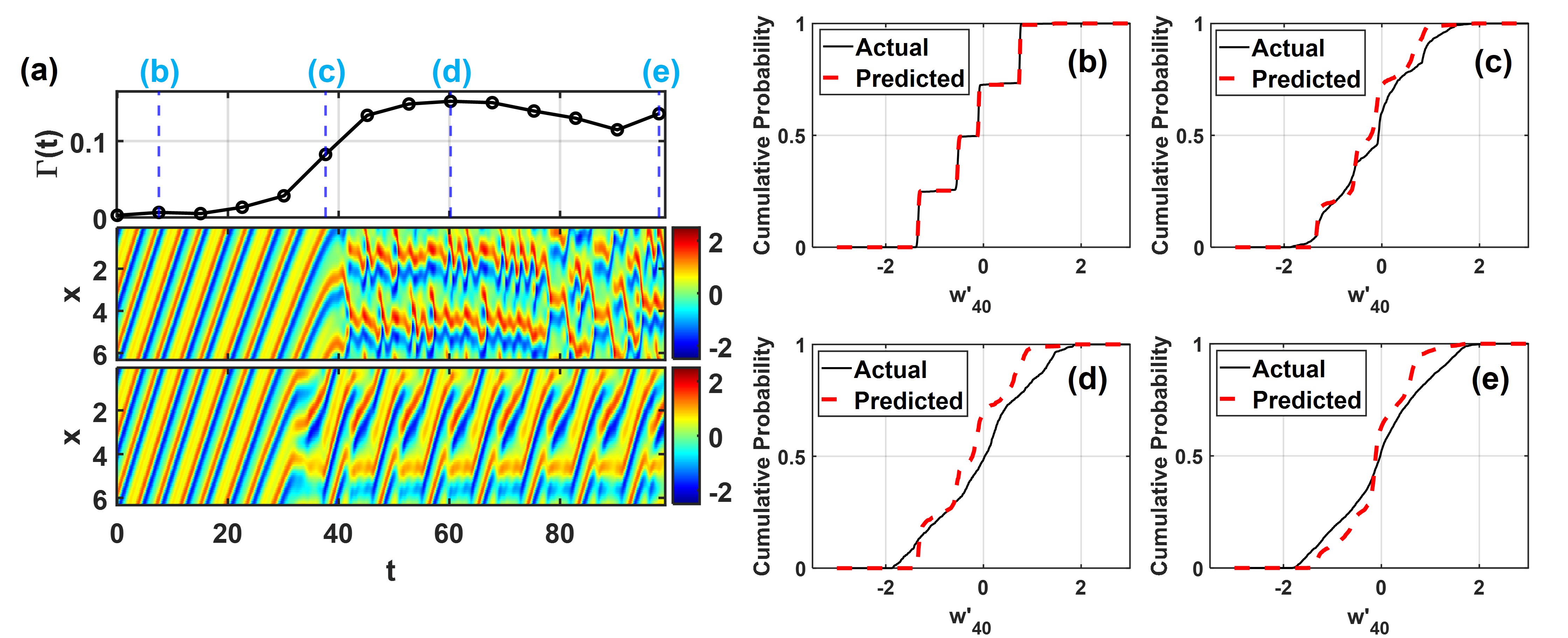}
    \caption{(a) (Top panel) $\Gamma(t)$ over the prediction window. Vertical dashed blue lines labeled (b), (c), (d), and (e) indicate the locations for which the true and predicted $w'_{40}$-Poincare map point cumulative probability distributions are shown in panels (b), (c), (d), and (e), respectively. (Middle Panel) An example of a typical numerically integrated system trajectory. (Bottom panel) The ML-prediction corresponding to the example trajectory in the middle panel. (b), (c), (d), and (e) show the true (solid black curves) and ML-predicted (dashed red curves) $w'_{40}$-Poincare map point cumulative probability distributions at $t=7.51,37.64,60.23$, and $97.88$, respectively.}
    \label{fig:ks_noiseless}
\end{figure}

We first consider the noiseless Kuramoto-Sivashinsky equation, i.e., $\xi = 0$ in Eq. (\ref{eqn:ks}), with the non-stationarity induced by the following time-dependence of $\kappa(t)$,

\begin{equation}
    \label{eqn:ks_kappa}
    \kappa(t) = \kappa_0 + \kappa_1 \exp{(t/\tau)},
\end{equation}

where $\kappa_0 = 0.0753$, $\kappa_1 = 0.0034$, and $\tau = 75.3$. Typical orbits of this non-stationary system evolve in periodic motion for $t<0$, and when $\kappa$ crosses a tipping point near $\kappa = 0.08$ they abruptly transition to chaotic motion. As before, our objective is to observe the system for $t<0$, optimize hyperparameters and train an ML model using this training data, and then predict the long-term time-evolution of the system trajectory for $t>0$. Figure \ref{fig:ks_noiseless} shows the results of applying our ML methods to this example using an ensemble of $1000$ trajectories from randomly chosen initial conditions. Figure \ref{fig:ks_noiseless}(a) shows the climate error metric $\Gamma(t)$ over the prediction window (the top panel), an example of a numerically integrated trajectory of the system, and the corresponding ML prediction (the middle panel and bottom panel, respectively) with the value of $w'(x,t)$ color-coded (where $w'(x,t)$ is $w(x,t)$ normalized by its root-mean-square with respect to time over the training data). The vertical dashed blue lines labeled (b), (c), (d), and (e) indicate the times for which the true and ML-predicted cumulative probability distributions of $w'_{40}$-Poincare map points are plotted in the corresponding panels of the figure. We see that initially the ML-predicted trajectories evolve in a periodic manner, in agreement with true system trajectories as indicated by the small value of $\Gamma$ for $t\lessapprox20$ and by the cumulative distributions in Fig. \ref{fig:ks_noiseless}(b). The ensemble of true trajectories undergo a tipping-point-transition from periodic to chaotic motion near $t=40$. We see from the example of an ML-predicted trajectory in the bottom panel in Fig. \ref{fig:ks_noiseless}(a) and from the cumulative distributions in Figs. \ref{fig:ks_noiseless}(c), \ref{fig:ks_noiseless}(d), and \ref{fig:ks_noiseless}(e) at $t=37.64, 64.23$, and $97.88$, respectively, that the ML-predicted trajectories undergo a transition from periodic motion to a different motion which does not match the chaotic motion of the true system. Thus, in this noiseless example, although the ML was able to anticipate a tipping point, it was not able to accurately capture the post-tipping-point transition dynamics.

 (Note, the cumulative probability distribution in Fig. \ref{fig:ks_noiseless}(b), at which point the system orbit is periodic, appears to indicate a period-$4$ motion, seemingly in disagreement with the bifurcation diagram in Fig. \ref{fig:ks_bifur} which shows a period-$2$ motion. This, however, is because the ensemble of trajectories used to calculate the statistics in Fig. \ref{fig:ks_noiseless} are initiated from randomly selected initial conditions, and because the wave propagation direction of the traveling wave solution in the pre-tipping-point regime has a symmetry, i.e., it can travel either from left to right or from right to left, the ensemble consists of $50\%$ of trajectories with waves traveling in one direction and $50\%$ traveling in the other direction. The $w'_{40}$-Poincare map points for each type of solution have different values and so the cumulative probability distribution generated from these map points shows $4$ steps instead of just the $2$.)
 
\begin{figure}[h]
    \centering
    \includegraphics[scale=0.45]{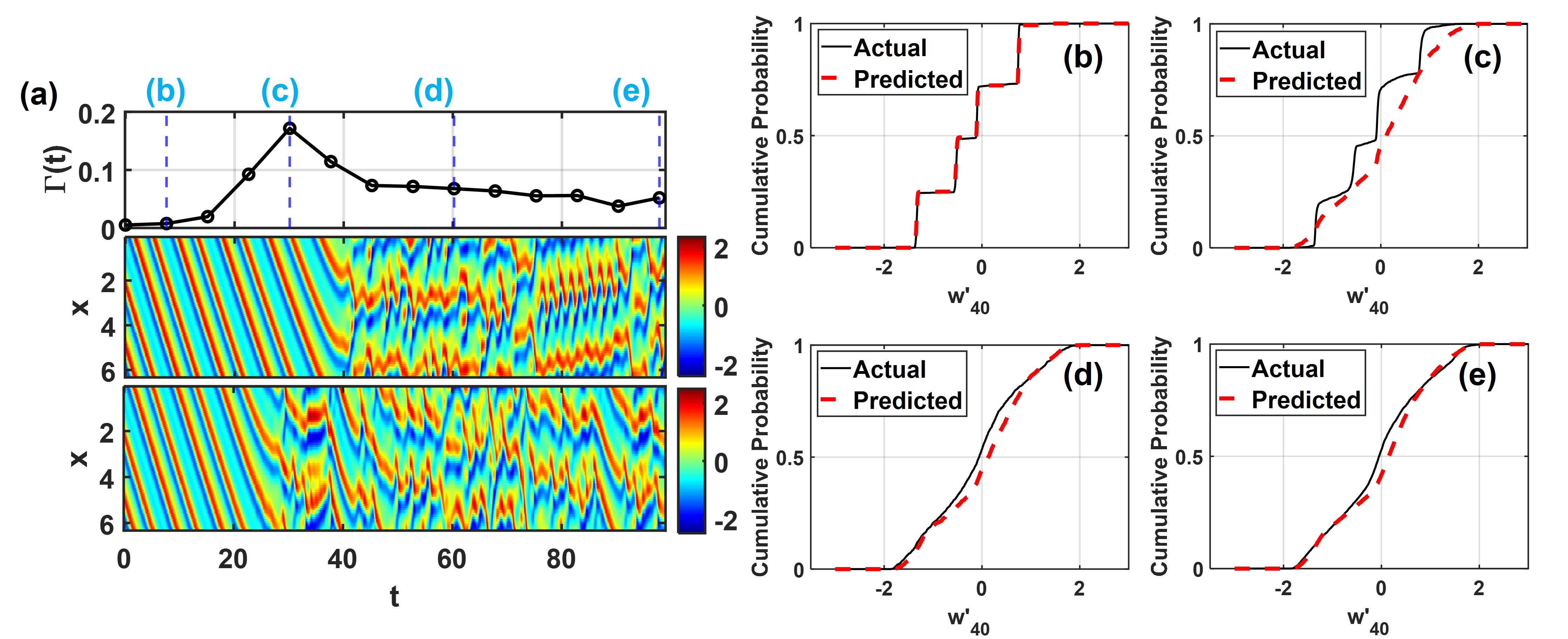}
    \caption{(a) (Top panel) $\Gamma(t)$ over the prediction window. Vertical dashed blue lines labeled (b), (c), (d), and (e) indicate the locations for which the true and predicted $w'_{40}$-Poincare map point cumulative probability distributions are shown in panels (b), (c), (d), and (e), respectively. (Middle Panel) An example of a typical numerically integrated system trajectory. (Bottom panel) The ML-prediction corresponding to the example trajectory in the middle panel. (b), (c), (d), and (e) show the true (solid black curves) and ML-predicted (dashed red curves) $w'_{40}$-Poincare map point cumulative probability distributions at $t=7.51,30.10,60.23$, and $97.88$, respectively.}
    \label{fig:ks_noisy}
\end{figure}

 As demonstrated with the Lorenz system in the previous section, we find that the presence of a small amount of dynamical noise can improve prediction, and may allow the ML to both anticipate the tipping point as well as accurately predict the post-tipping-point dynamics. Thus, we reconsider the above system, but now evolving in the presence of dynamical noise. We implement the dynamical noise by adding a random number $\xi_i$ for $i=1,2,...64$ to our numerical solution $w(x_i,t)$ at spatial grid point $x_i$ at each iteration, where the random number, independent for each spatial grid point, is chosen randomly from a uniform distribution over the range $[-1\times 10^{-4}, 1\times 10^{-4}]$. Figure \ref{fig:ks_noisy} shows the result of applying our ML methods to this example for an ensemble of $1000$ trajectories from randomly chosen initial conditions. Figure \ref{fig:ks_noisy}(a) shows $\Gamma(t)$ over the prediction window, an example of a numerically integrated system trajectory, and the corresponding ML-prediction in the top, middle, and bottom panels, respectively. Compared to $\Gamma(t)$ in the noiseless case in Fig. \ref{fig:ks_noiseless}(a), we see that $\Gamma(t)$ in this case remains small throughout the prediction window. The example true and ML-predicted trajectories further show that the ML-predicted trajectory successfully anticipates the tipping point and qualitatively captures the post-tipping-point dynamics. This is quantitatively supported by the plots of the true and ML-predicted cumulative probability distributions in Figs. \ref{fig:ks_noisy}(b), \ref{fig:ks_noisy}(c),\ref{fig:ks_noisy}(d), and \ref{fig:ks_noisy}(e), taken at times $t=7.51,30.10,60.23$, and $97.88$, respectively. We  see that initially the ML-predicted ensemble of trajectories are periodic and then transition to a chaotic state which closely approximates that of the true system. There is a spike in $\Gamma(t)$ at $t=30.10$, and from the cumulative probability distributions in Fig. \ref{fig:ks_noisy}(b) at that time, we see that this spike is due to a discrepancy in the timing of the tipping-point transition in the ML-predictions compared to in the true system orbits. This peak quickly reduces, however, for later times where the ensembles of true and predicted trajectories have both undergone a tipping-point transition from periodic to chaotic. We see that, while in the noiseless case the ML was unable to predict the post-tipping-point transition dynamics, the presence of a small amount of dynamical noise allowed the ML to predict those dynamics.
 
 To enable accurate climate prediction of the high-dimensional KS system in this section, we used the first modification to the reservoir computing architecture introduced in Appendix B, i.e., we separated the input coupling of the dynamical variables (i.e., $w(x,t)$) and the linear control signal. Table \ref{tab:ks} shows the hyperparameters used in the Kuramoto-Sivashinsky equation examples.

\begin{table}[h]
    \centering
    \caption{Hyperparameters for predicting the Kuramoto-Sivashinsky equation.}
    \begin{tabular}{m{25em}|m{4em}|m{6em}}
         Reservoir size & $N$ & $3000$  \\
         \hline
         Average number of connections per node & $\langle d \rangle$ & $3$\\
         \hline
         Spectral radius of the reservoir's adjacency matrix & $\rho_s$ & $1$\\
         \hline
         Maximal coupling strength of the dynamical variable input to the reservoir & $\chi_1$ & $1$\\
         \hline
         Maximal coupling strength of the parameter input to the reservoir & $\chi_2$ & $0.75$\\
         \hline
         Reservoir leakage parameter & $\alpha$ & $0.5$\\
         \hline
         Reservoir node activation bias & $b$ & $1$\\
         \hline
         Tikhonov regularization parameter & $\alpha$ & $7.74\times 10^{-10}$\\
         \hline
         Intercept of the linear control signal & $b$ & $1$\\
         \hline
         Slope of the linear control signal & $a$ & $1\times 10^{-5}$\\
         \hline
         Numerical integration time step for Eq. (\ref{eqn:ks}) & $\Delta t$ & $0.0084$\\
         \hline
         RC time step & $\Delta t_{RC}$ & $0.0084$\\
         \hline
         Strength of observational noise added to training data & $\epsilon_0$ & $0$\\
         \hline
         Number of passes of training data during training & &$1$\\
    \end{tabular}
    \label{tab:ks}
\end{table}

\subsection{Hysteretic tipping in the Lorenz System: Hybrid ML/knowledge-based prediction}

We have demonstrated through the previous examples how ML is, in many cases, capable of predicting non-hysteretic tipping point dynamics. We also discussed limitations on the ML's ability to extrapolate (e.g., when the system non-stationarity is too slow and the training data is too short (Sec. 3.1)), as well as how the presence of dynamical noise in the system may help the ML to successfully extrapolate the post-tipping-point dynamics. Here we provide another example illustrating the limitation of the method, particularly for cases where a tipping point is mediated by a hysteretic bifurcation of the corresponding stationary system, and we illustrate how using a hybrid system combining an ML model with even an inaccurate, conventional knowledge-based model may help overcome the limitation. The example is that of a tipping point in the non-stationary Lorenz `63 system associated with a subcritical (i.e., hysteretic) Hopf bifurcation. 

\begin{figure}[h]
    \centering
    \includegraphics[scale=0.75]{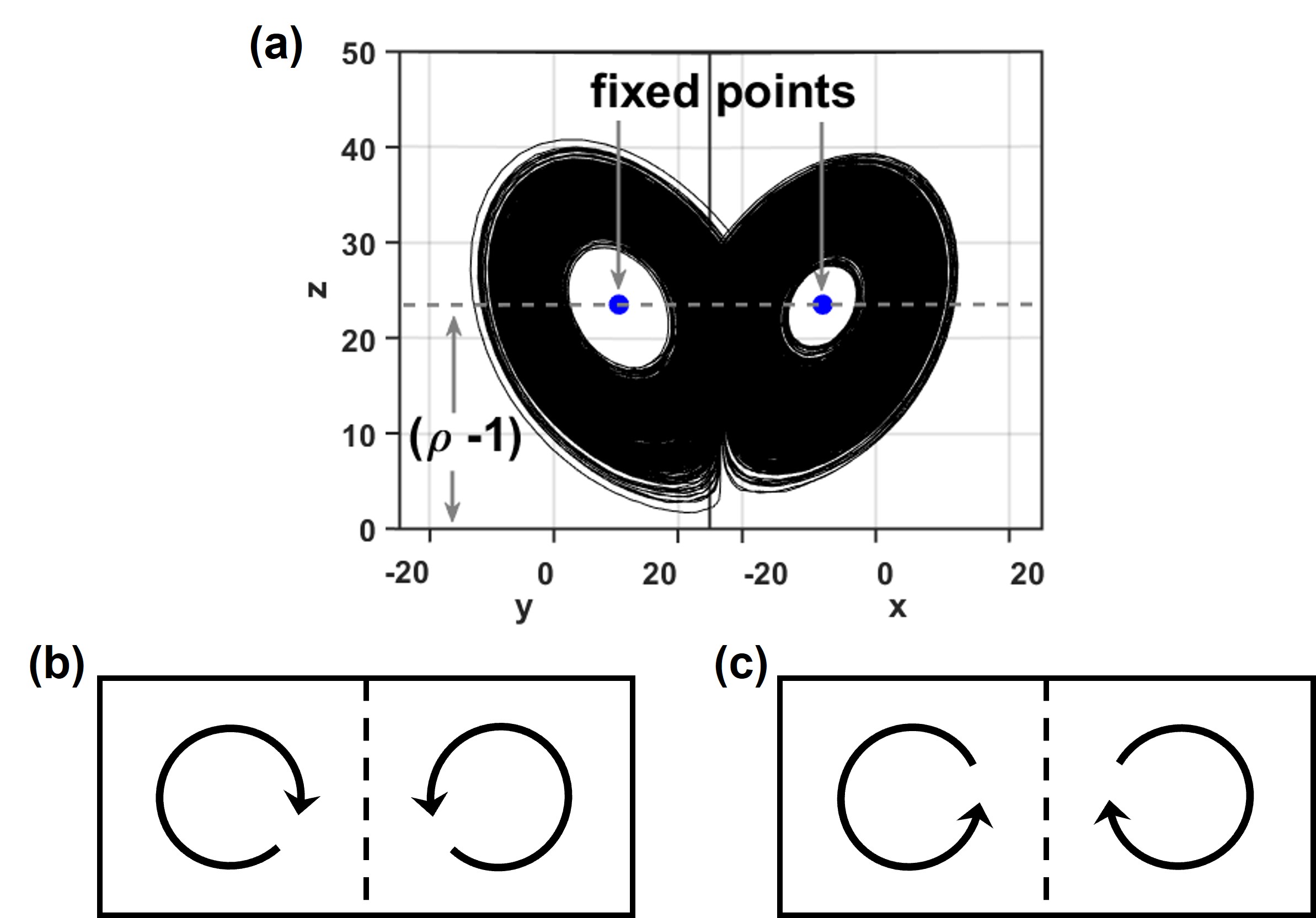}
    \caption{(a) For $\rho$ slightly less than the critical bifurcation value an attracting chaotic orbit of  the  Lorenz system (black curve) coexists with two fixed point attractors (blue dots) (representing the convective fluid flow patterns of the Lorenz  model in panels (b) and (c)).}
    \label{fig:lorenz_sh_stationary}
\end{figure}

A system undergoes a Hopf bifurcation when a fixed point attractor loses stability as a pair of complex conjugate eigenvalues of its linearized system cross the imaginary axis. The bifurcation is subcritical when there exists an unstable limit cycle about the pre-bifurcation stable fixed point which collapses onto the fixed point as the bifurcation is approached, and past which the periodic orbit becomes unstable. In such a case the post-bifurcation orbit will be expelled from the neighborhood of the pre-bifurcation fixed point attractor, typically moving to a previously existing attractor, possibly located in a far away region of state space. For the Lorenz system in our example the post-bifurcation attractor is chaotic (see Ref. \cite{yorke} and Section $8.4$ of Ref. \cite{ott} for details). In our considered example, for the $t<0$ training data, the system moves along one of two slowly-changing fixed point attractors symmetrically located in the central regions of the left and right ``butterfly wings" of the Lorenz attractor ($x=y=\pm\sqrt{\beta(\rho -1)}, z=\rho -1$). See Fig. \ref{fig:lorenz_sh_stationary}. At a time after the start of the prediction phase ($t>0, \rho(0) = \rho_0 + \rho_1)$, the system undergoes a tipping point transition mediated by a subcritical Hopf bifurcation of the corresponding stationary system. (Because of the Lorenz system symmetry with respect to the transformation $(x, y, z) \rightarrow (-x, -y, z)$, the Hopf bifurcation of both of the fixed points occur simultaneously.)

\begin{figure}[h]
    \centering
    \includegraphics[scale=0.7]{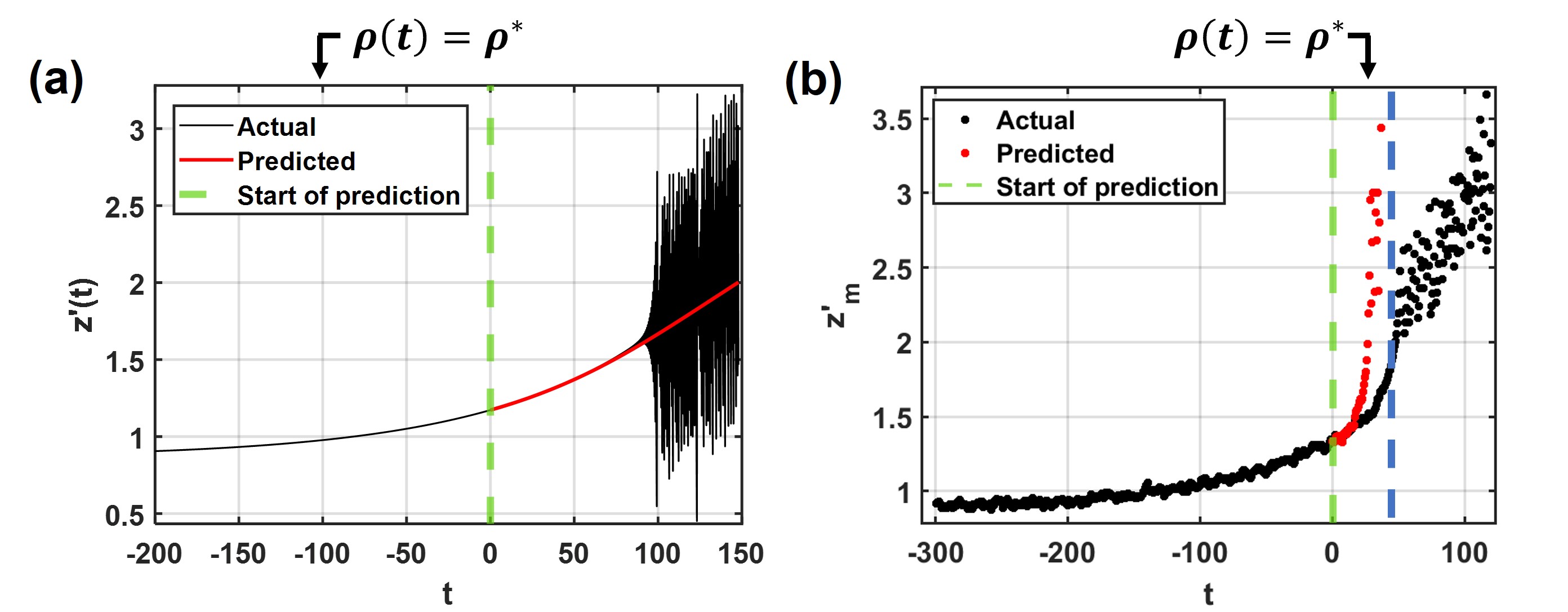}
    \caption{(a) $z'(t)$ of an example true system orbit (black curve) and of the corresponding ML prediction (red curve) for the case of the noiseless Lorenz system. The vertical green dashed line indicates the start of the ML prediction. To describe panel (b), we first define the quantity $z'_m$ as $z$-coordinate values that occur when the noisy trajectory crosses the surface $x = \beta z$. This surface corresponds to the surface of section of the chaotic noiseless system at local $z(t)$ maxima of the noiseless trajectory, which, from Eq. \ref{eqn:lorenz}, occur at $dz/dt = xy - \beta z = 0$. Panel (b) shows the $z'_m$ values of an example true noisy Lorenz system orbit (black dots) versus $t$ and of the corresponding ML prediction (red dots). The vertical green dashed line marks the start of the ML prediction, and the vertical blue dashed line marks the time near which the ML-predicted trajectory leaves the plotted range due to instability of the ML prediction dynamics. Note that, in these panels, the two noiseless fixed points ($x=y=\pm\sqrt{\beta(\rho -1)}, z=\rho -1$) have the same $z$-coordinate and would appear as a single curve.}
    \label{fig:lorenz_sh}
\end{figure}

First considering the noiseless case, we set $\rho_0=22$, $\rho_1=7.5$, and $\tau=100$. Figure \ref{fig:lorenz_sh}(a) shows the $z'(t)$-component of an example true system trajectory and the corresponding ML prediction in black and red curves, respectively. The vertical green dashed line shows the starting point for the ML prediction. We see that the true system orbit (plotted in black) initially moves along a slowly-changing fixed point of the system, and subsequently abruptly transitions to a chaotic motion. The ML-predicted-orbit (plotted in red), on the other hand, continues to evolve along the slowly-changing fixed point even after the true system fixed point has become unstable. Note that in this example the bifurcation point of the corresponding stationary system ($\rho=\rho^*=24.74$) is crossed well before the tipping process commences in the non-stationary system (as illustrated in Fig. \ref{fig:lorenz_sh}(a)). Table \ref{tab:lorenz_sh_noiseless} lists the hyperparameters used for this example.

\begin{table}[h]
    \centering
    \caption{Hyperparameters for predicting the noiseless Lorenz system (subcritical Hopf bifurcation).}
    \begin{tabular}{m{25em}|m{4em}|m{6em}}
         Reservoir size & $N$ & $2000$  \\
         \hline
         Average number of connections per node & $\langle d \rangle$ & $3$\\
         \hline
         Spectral radius of the reservoir's adjacency matrix & $\rho_s$ & $1$\\
         \hline
         Maximal coupling strength of the input to the reservoir & $\chi$ & $0.75$\\
         \hline
         Reservoir leakage parameter & $\alpha$ & $0.5$\\
         \hline
         Reservoir node activation bias & $b$ & $0$\\
         \hline
         Tikhonov regularization parameter & $\alpha$ & $1.29\times 10^{-12}$\\
         \hline
         Intercept of the linear control signal & $b$ & $1$\\
         \hline
         Slope of the linear control signal & $a$ & $3.59\times 10^{-5}$\\
         \hline
         Numerical integration time step for Eq. (\ref{eqn:lorenz}) & $\Delta t$ & $0.01$\\
         \hline
         RC time step & $\Delta t_{RC}$ & $0.01$\\
         \hline
         Strength of observational noise added to training data & $\epsilon_0$ & $1\times 10^{-4}$\\
         \hline
         Number of passes of training data during training & &$10$\\
    \end{tabular}
    \label{tab:lorenz_sh_noiseless}
\end{table}

For the noisy case we set $\rho_0=15$, $\rho_1=7.5$, and $\tau=100$. Figure \ref{fig:lorenz_sh}(b) shows $z'_m$ points versus time for a typical true system trajectory (black dots) and the corresponding ML-predicted trajectory (red dots) for a case where the ML model is trained on data corresponding to $t\in[-300, 0]$ and prediction begins at $t=0$. For this example, the $z'_m$ points are the values of $z$ (normalized by the root-mean-square of $z(t)$ over the training data) where $xy - \beta z = 0$ (see figure caption for additional description). We see that, in agreement with the true system trajectory, the ML-predicted trajectory initially evolves noisily about the slowly-changing stable fixed point and then predicts that the fixed point becomes unstable at a later time. However, while the post-tipping-point trajectory of the true system evolves on the bounded chaotic attractor, the ML-predicted trajectory quickly diverges to regions very far from the now unstable fixed point, indicating that the ML model is unable to successfully extrapolate the system dynamics to the post-tipping-point chaotic set. The vertical dashed blue line indicates the time near which the ML-predicted trajectory leaves the range of plotted values. This example illustrates a case in which our proposed ML method, although successful at predicting the occurrence of a tipping point, is unable to extrapolate to post-tipping-point dynamics presumably because the information available from the training data is limited to a region that is too far removed from the region where the post-tipping dynamics occurs. Table \ref{tab:lorenz_sh_noisy} lists the hyperparameters used in this example.

\begin{table}[h]
    \centering
    \caption{Hyperparameters for predicting the noisy Lorenz system (subcritical Hopf bifurcation).}
    \begin{tabular}{m{25em}|m{4em}|m{6em}}
         Reservoir size & $N$ & $2000$  \\
         \hline
         Average number of connections per node & $\langle d \rangle$ & $3$\\
         \hline
         Spectral radius of the reservoir's adjacency matrix & $\rho_s$ & $0.25$\\
         \hline
         Maximal coupling strength of the input to the reservoir & $\chi$ & $0.25$\\
         \hline
         Reservoir leakage parameter & $\alpha$ & $1$\\
         \hline
         Reservoir node activation bias & $b$ & $0$\\
         \hline
         Tikhonov regularization parameter & $\alpha$ & $1\times 10^{-13}$\\
         \hline
         Intercept of the linear control signal & $b$ & $1$\\
         \hline
         Slope of the linear control signal & $a$ & $1.29\times 10^{-5}$\\
         \hline
         Numerical integration time step for Eq. (\ref{eqn:lorenz}) & $\Delta t$ & $0.01$\\
         \hline
         RC time step & $\Delta t_{RC}$ & $0.01$\\
         \hline
         Strength of observational noise added to training data & $\epsilon_0$ & $1\times 10^{-5}$\\
         \hline
         Number of passes of training data during training & &$10$\\
    \end{tabular}
    \label{tab:lorenz_sh_noisy}
\end{table}

\begin{figure}[h]
    \centering
    \includegraphics[scale=0.5]{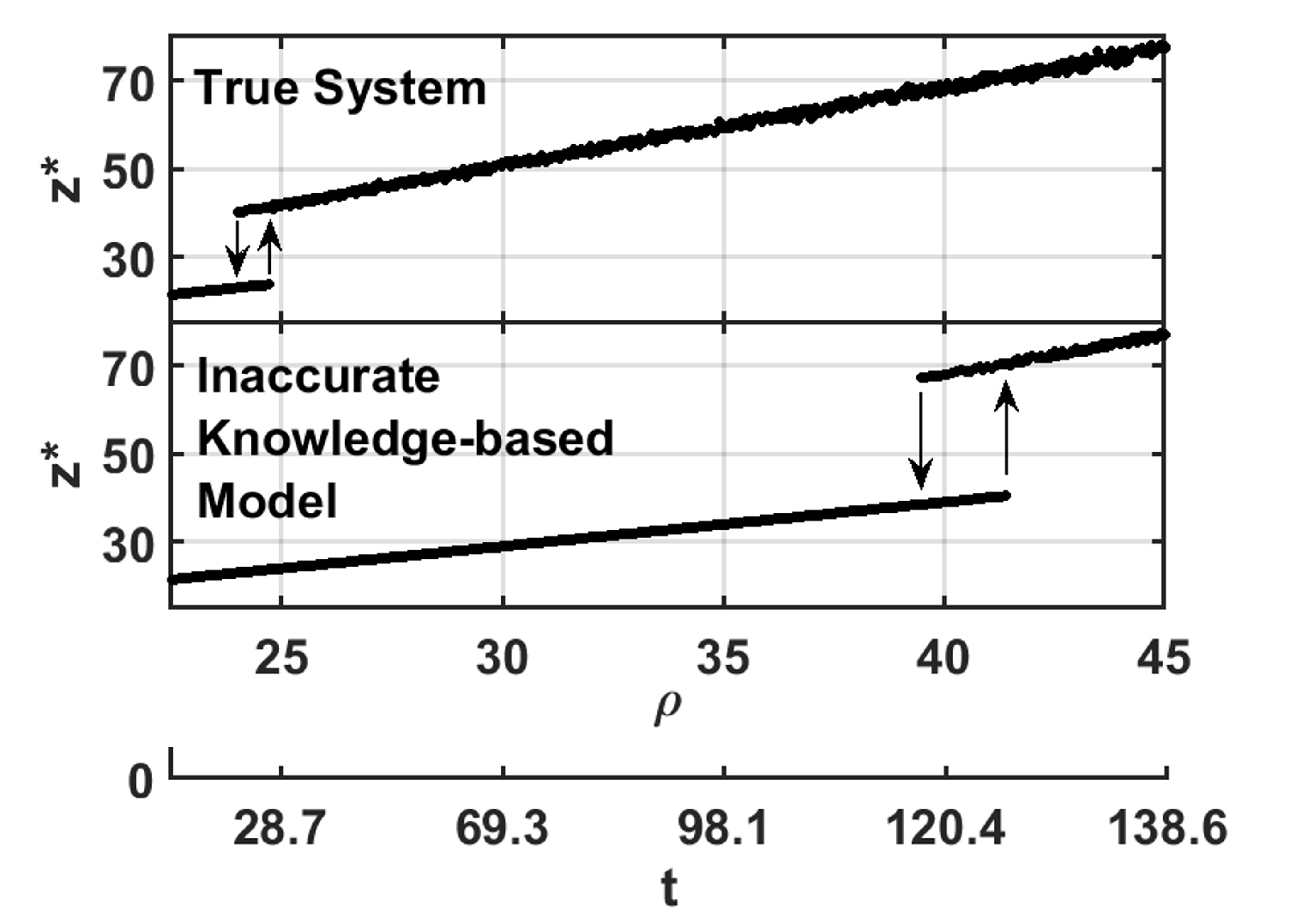}
    \caption{Stationary bifurcation diagrams showing the hysteretic subcritical Hopf bifurcation for the true noiseless Lorenz system (top panel) and the inaccurate knowledge-based model (bottom panel) for $\rho \in [22.5, 45]$, where the bifurcation plots are constructed by plotting $z^*$ against $\rho$. For fixed values of $\rho$, $z^*$ are either the maximum $z$ value over long trajectories (for orbits on the chaotic attractor) or the $z$ value for the fixed point attractor. The bottom horizontal axis labeled $t$ gives the times at which the non-stationary system $\rho$ value passes through the $\rho$ values on the $\rho$ axis immediately above this $t$ axis.}
    \label{fig:lorenz_sh_hysteresis_loop}
\end{figure}

\begin{figure}[h]
    \centering
    \includegraphics[scale=0.45]{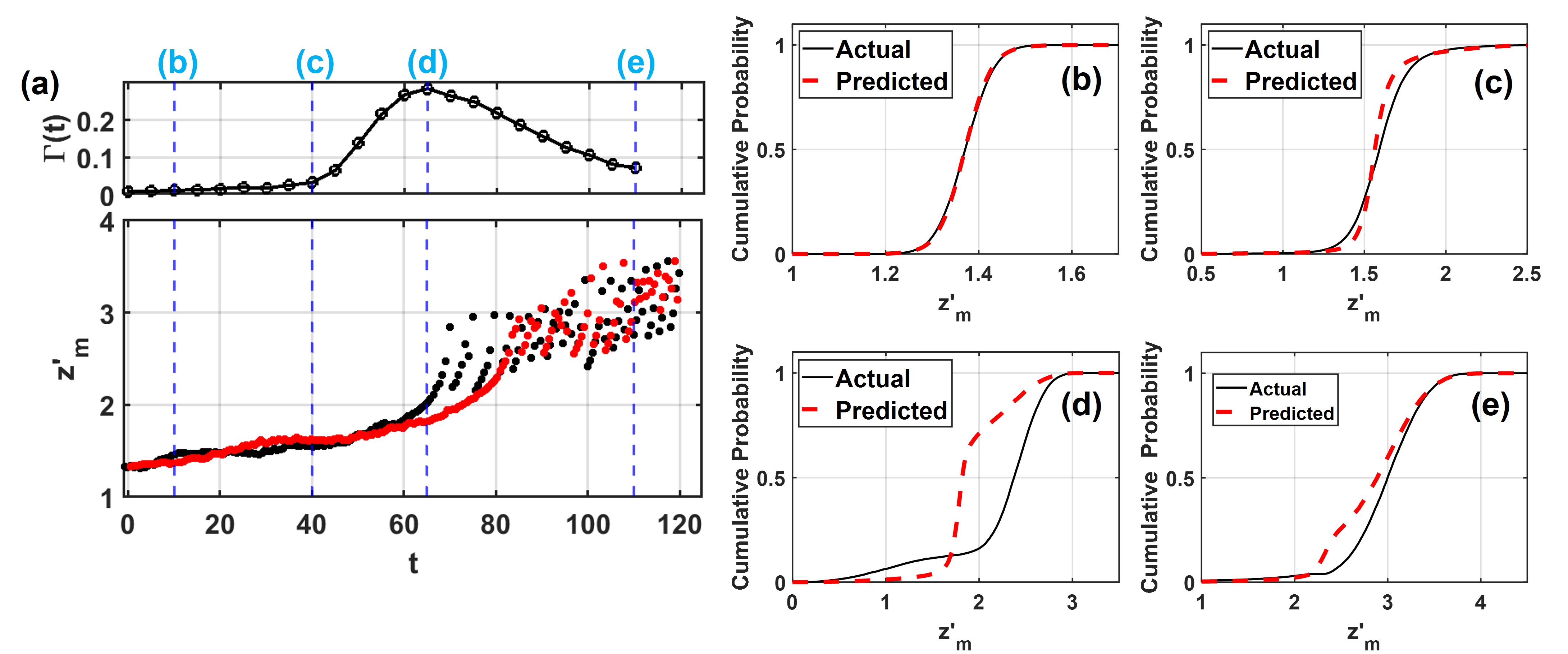}
    \caption{(a) (Top panel) $\Gamma(t)$ over the prediction window. Vertical dashed blue lines labeled (b), (c), (d), and (e) indicate the locations for which the true and predicted $z'_m$ point cumulative probability distributions are shown in panels (b), (c), (d), and (e), respectively. (See caption of Fig. \ref{fig:lorenz_sh}(b) for the definition of $z'_m$.) (Bottom Panel) The $z'_m$ points of a typical true system trajectory (black dots), and the corresponding ML/knowledge-based model prediction (red dots). (b), (c), (d), and (e) show the true (solid black curves) and ML/knowledge-based model predicted (dashed red curves) $z'_m$ point cumulative probability distributions at $t=10,40,65$, and $110$, respectively.}
    \label{fig:lorenz_sh_hybrid}
\end{figure}

As we saw above there are tipping point cases which we may be interested in predicting, but due to the nature of the tipping point (e.g., mediated by a hysteretic bifurcation of the corresponding stationary system) a purely data-driven ML approach can fail to successfully predict the post-tipping-point system dynamics (e.g., due to the fact that the post-tipping-point system orbit explores regions of state space which may be too far from the regions explored by pre-tipping-point training orbit). In such cases, additional valuable information about the regions of the state space to which the data-driven ML model fails to extrapolate can be provided by a knowledge-based model. Next we consider the same example as above, but we replace the purely data-driven ML setup with a hybrid ML setup which incorporates an inaccurate knowledge-based physics model. We show that a hybrid model which combines a data-driven component and a knowledge-based component, neither of which can individually make useful predictions, is able to predict the occurrence of a tipping point, as well as the associated post-tipping-point dynamics. The hybrid ML model architecture and training are described in Appendix C. Here we use the noiseless non-stationary Lorenz system with the same $\rho(t)$ as the true system, but with $\beta=16/3$ and $\sigma=20$ (compared to $\beta=8/3$ and $\sigma=10$ in the true system) as our inaccurate knowledge-based model. Figure \ref{fig:lorenz_sh_hysteresis_loop} shows stationary system bifurcation diagrams of the true noiseless Lorenz model (top panel) and the inaccurate knowledge-based model (bottom panel), plotted as $z^*$ versus $\rho$. Here $z^*$ is defined as follows: the lower curve in both panels of Fig. \ref{fig:lorenz_sh_hysteresis_loop} corresponds to a fixed point attractor, and for this curve we take $z^*$ to denote the $z$-coordinate of the fixed point; the upper curve in both panels of Fig. \ref{fig:lorenz_sh_hysteresis_loop} corresponds to a chaotic attractor and for this curve we take $z^*$ to denote the maximum $z_m$ over a long trajectory of the system. The upward-pointing arrow in each panel indicates the hysteretic transition from the fixed point attractor to the chaotic attractor as $\rho$ is increased. The downward-pointing arrow in each panel indicates the hysteretic transition from the chaotic attractor to the fixed point attractor as $\rho$ is decreased. The bottom horizontal axis labeled $t$ shows the time corresponding to the value of $\rho$ for the non-stationary system example $\rho = \rho(t)$ (see Fig. \ref{fig:lorenz_sh_hybrid}). We note that the critical bifurcation value of $\rho$ for the inaccurate system is substantially above the critical bifurcation value for the true system, and the region of bistability of the inaccurate knowledge-based model begins near the right edge of the prediction window for this example (near $t = 116$).

Figure \ref{fig:lorenz_sh_hybrid} shows the result of applying our hybrid ML/knowledge-based model to the above case using an ensemble of $2000$ trajectories from randomly chosen initial conditions. The top panel of Fig. \ref{fig:lorenz_sh_hybrid}(a) shows $\Gamma(t)$ over the prediction window and the bottom panel shows the $z'_m$ points versus time for a typical true system trajectory (black dots) and the corresponding ML prediction (red dots). The vertical dashed blue lines at $t=10,40,65$, and $110$ indicate the times for which the $z'_m$ point cumulative probability distributions are plotted in Figs. \ref{fig:lorenz_sh_hybrid}(b),\ref{fig:lorenz_sh_hybrid}(c), \ref{fig:lorenz_sh_hybrid}(d), and \ref{fig:lorenz_sh_hybrid}(e), respectively. We see from the bottom panel of Fig. \ref{fig:lorenz_sh_hybrid}(a) that the ML-predicted trajectory initially evolves noisily about the slowly drifting fixed point attractor of the corresponding stationary system, and then undergoes a tipping point, after which it begins to explore a much larger region of state space, in agreement with the true system orbit. Unlike predictions using the purely data-driven ML model (see Fig. \ref{fig:lorenz_sh}(b)), the predicted hybrid ML/knowledge-based post-tipping-point trajectories remain bounded and exhibit dynamics similar to those exhibited by the true system trajectories. As before, we see that although the predicted hybrid ML/knowledge-based trajectories capture the tipping point and post-tipping-point dynamics well, they do not accurately predict the timing of the tipping point. This is seen in the rise in $\Gamma(t)$ near $t=65$, at which point most of the true system trajectories have undergone a tipping point, but many of the predicted hybrid ML/knowledge-based trajectories have not. This spike in $\Gamma(t)$ is  reduced by the end of the prediction window, since by that point almost all of the ML-predicted trajectories have undergone a tipping point transition (we found that less than $1\%$ of the ensemble of trajectories predicted by the hybrid ML/knowledge-based system failed to undergo a tipping point). We see in this example that although the purely data-driven ML model and the knowledge-based physics model were individually unable to make useful predictions, a hybrid model which combined the two was able to successfully predict the tipping point and post-tipping-point dynamics. We note that, in general, a more (or less) accurate knowledge-based component would allow for better (or worse) performance of the hybrid model. 

\begin{table}[h]
    \centering
    \caption{Hyperparameters for predicting the noisy Lorenz system (subcritical Hopf bifurcation) using the hybrid ML model.}
    \begin{tabular}{m{25em}|m{4em}|m{6em}}
         Reservoir size & $N$ & $2000$  \\
         \hline
         Average number of connections per node & $\langle d \rangle$ & $3$\\
         \hline
         Spectral radius of the reservoir's adjacency matrix & $\rho_s$ & $0.25$\\
         \hline
         Maximal coupling strength of the input to the reservoir & $\chi$ & $0.1$\\
         \hline
         Reservoir leakage parameter & $\alpha$ & $1$\\
         \hline
         Reservoir node activation bias & $b$ & $0$\\
         \hline
         Tikhonov regularization parameter & $\alpha$ & $1\times 10^{-8}$\\
         \hline
         Intercept of the linear control signal & $b$ & $1$\\
         \hline
         Slope of the linear control signal & $a$ & $1.29\times 10^{-5}$\\
         \hline
         Numerical integration time step for Eq. (\ref{eqn:lorenz}) & $\Delta t$ & $0.01$\\
         \hline
         RC time step & $\Delta t_{RC}$ & $0.01$\\
         \hline
         Strength of observational noise added to training data & $\epsilon_0$ & $0$\\
         \hline
         Number of passes of training data during training & &$1$\\
    \end{tabular}
    \label{tab:lorenz_sh_noisy_hybrid}
\end{table}

\section{Conclusion}

In this paper, we have addressed the problem of using machine learning to predict the tipping point transition of non-stationary dynamical systems that exhibit a constrained pre-tipping-point motion which is restricted to a small subset of its system state space (e.g., a periodic orbit), and which, upon crossing a tipping point, abruptly transitions to a chaotic motion that evolves on a larger region of its system state space that was unexplored, or only  sparsely explored, during its pre-tipping-point motion used for training. In particular, we demonstrated that a ML model trained on the pre-tipping-point orbit of a non-stationary dynamical system can, under some circumstances, usefully predict a tipping point as well as chaotic post-tipping-point behavior. We explored limitations to our methods (e.g., system non-stationarity is too slow, or the system tipping point is mediated by a hysteretic bifurcation of the corresponding stationary system), and possible scenarios and approaches (e.g., the system is evolving in the presence of dynamical noise, or using a hybrid ML/knowledge-based scheme). We demonstrated this using the Lorenz `63 system, the Ikeda map, and the Kuramoto-Sivashinsky equation with non-stationarity induced by a time-dependent system parameter drift (of which we assumed no knowledge). A key element in enabling successful prediction in these examples is the new hyperparameter optimization scheme introduced in Sec. 2.2. The main conclusions of our paper are as follows: 

\begin{enumerate}
    \item ML is a promising tool for learning the dynamics of a non-stationary dynamical system from the time series of past system states, anticipating potential tipping points in the future, and predicting post-tipping-point behavior on state space regions not visited, or only sparsely visited, by the training data.
    \item As with all extrapolation methods, the methods presented in this paper may fail when the ``amount" of extrapolation from the known case increases too much. In a few such cases we found that
    \begin{enumerate}
        \item the presence of dynamical noise in the system can play a beneficial role in enabling the ML to make useful predictions.
        \item a hybrid ML/knowledge-based model, where the ML and knowledge-based components individually fail to any make useful predictions, can predict future tipping points and post-tipping-point behavior of the system.
    \end{enumerate}
    \item Obtaining good predictions for non-stationary dynamical systems may require careful tuning of the model hyperparameters, and we propose a method for choosing an appropriate set of hyperparameters which we found useful for the test examples considered in this paper, and that we believe may be generally useful for predicting the climate of any non-stationary dynamical system. 
\end{enumerate}

\section*{Appendix A: Choosing appropriate hyperparameters for non-stationary systems}

The performance of a ML model can be highly sensitive to the combination of the choice of hyperparameters used to create and train it, and the task the model is asked to perform. As such, an optimal set of hyperparameters is often chosen to be the set which allows a ML model, trained on a subset of the training data, to perform the target task the best on a disjoint subset of the training data not used during model training, called the validation data set. For forecasting problems, the model hyperparamters are typically optimized by maximizing a valid prediction time. Although this works well for predicting the short-term time-evolution of the state of the target system, we found that it does not always yield hypeparameters which allow the ML model to predict long-term trajectories which capture accurate climate dynamics including tipping-point transitions. For this reason we developed the hyperparameter optimization scheme for non-stationary systems described in Sec. 2.2 and used for the numerical experiments of this paper which incorporates a long validation data set to ensure that the ML predicted trajectories are long-term stable and accurately capture the climate dynamics of the system. Here we will, for the purpose of comparison, demonstrate the predictive performance of the ML model used in one of the examples in the paper, but where the model is created using hyperparameters obtained from maximizing only the median valid time over a short validation data set.

\begin{figure}[h]
    \centering
    \includegraphics[scale=0.45]{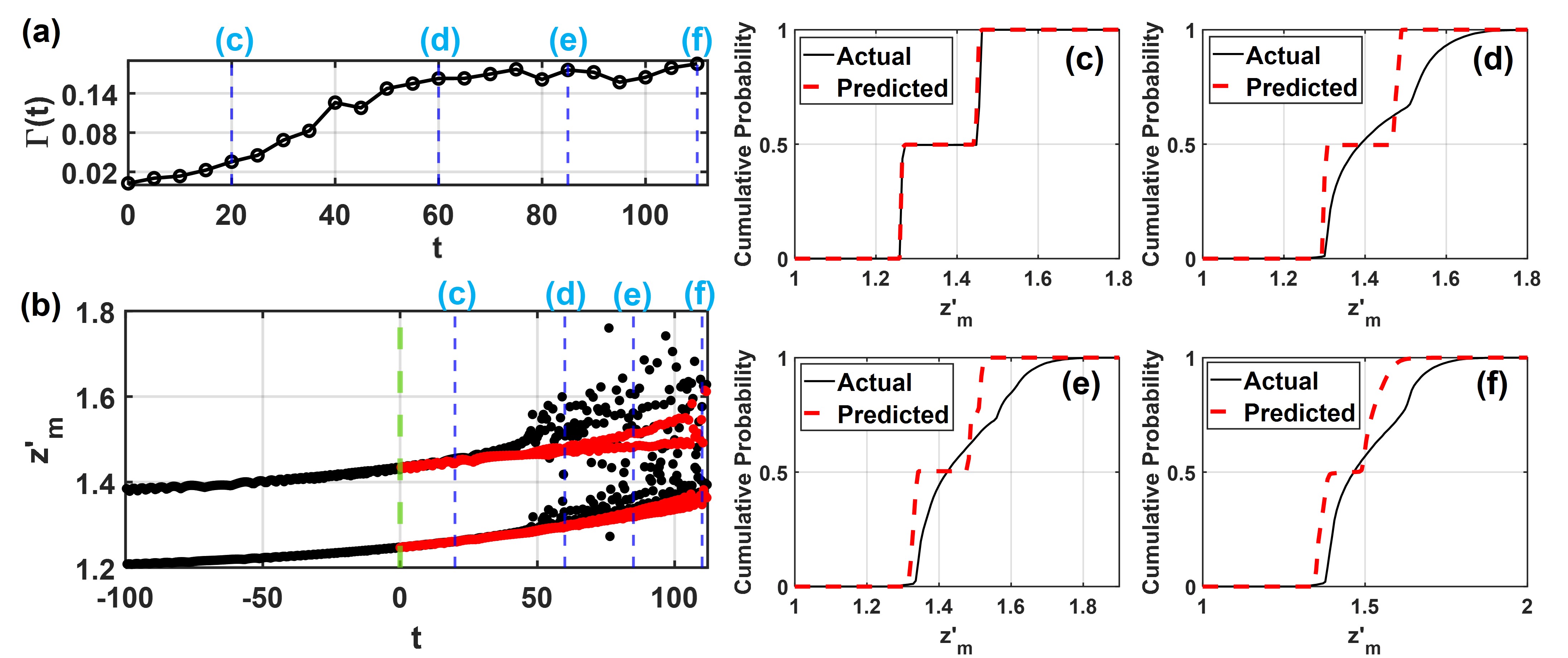}
    \caption{(a) $\Gamma(t)$ over the prediction window. Vertical dashed blue lines labeled (c), (d), (e), and (f) indicate the locations for which the true and predicted $z'_m$ point cumulative probability distributions are shown in panels (c), (d), (e), and (f), respectively. (b) The $z'_m$ points of an example true system trajectory (black dots) and the corresponding ML prediction (red dots). The vertical dashed green line indicates the starting point of the ML prediction. The vertical dashed blue lines indicate the same as in panel (a). (c), (d), (e), and (f) show the true (solid black curves) and ML-predicted (dashed red curves) $z'_m$ point cumulative probability distributions at $t=20,60,85$, and $110$, respectively.}
    \label{fig:lorenz_noiseless_tau100_hp_mvt}
\end{figure}

We here use the example of the non-stationary, noiseless Lorenz system (Eqs. (\ref{eqn:lorenz})) and the $\rho$-parameter drift given by Eq. (\ref{eqn:exp_param}) with $\rho_0=154$, $\rho_1=8$, and $\tau=100$. This is the first example considered in Sec. 3.1. Figure \ref{fig:lorenz_noiseless_tau100_hp_mvt} shows the result of applying our ML methods to an ensemble of $1000$ trajectories from randomly chosen initial conditions, but the ML model hyperparameters (listed in Table \ref{tab:lorenz_mvt_hp}) are optimized by \emph{only maximizing the median valid time}. Compared to the case in Fig. \ref{fig:lorenz_noiseless_tau100} where the hyperparameters used were obtained using our proposed scheme, we see that the climate error metric $\Gamma(t)$ in Fig. \ref{fig:lorenz_noiseless_tau100_hp_mvt}(a) is quite large for most of the prediction window. This is because the ML-predicted ensemble of trajectories fail to undergo a tipping point transition from periodic motion to chaotic motion, as seen from the $z'_m$ points cumulative probability distributions in Figs.\ref{fig:lorenz_noiseless_tau100_hp_mvt}(c), \ref{fig:lorenz_noiseless_tau100_hp_mvt}(d), \ref{fig:lorenz_noiseless_tau100_hp_mvt}(e),and \ref{fig:lorenz_noiseless_tau100_hp_mvt}(f) at times $t=20,60,85$, and $110$, respectively. From the true and ML-predicted $z'_m$ points of an example trajectory in Fig. \ref{fig:lorenz_noiseless_tau100_hp_mvt}(b), we see that the ML-predicted trajectories do not reliably predict the non-stationary climate dynamics of the target system. This example is meant to compare the quality of predictions of an ML model based on the hyperparameters that are used to create the model. Although it is a single example, and by no means a comprehensive study, it highlights the basic patterns we observed, namely, that determining hyperparameters by optimizing just the median valid time over a short validation data set did not consistently yield hyperparameters which enabled the ML model to reliably predict the non-stationary system climate evolution in the examples we studied.  In contrast, using the scheme proposed in Sec. 2.2 consistently allowed us to obtain hyperparameters which enabled the ML model to reliably predict the non-stationary climate evolution of the systems studied in this paper for various parameter settings.

\begin{table}[h]
    \centering
    \caption{Hyperparameters for predicting the Lorenz equations in Fig. \ref{fig:lorenz_noiseless_tau100_hp_mvt}}
    \begin{tabular}{m{25em}|m{4em}|m{6em}}
         Reservoir size & $N$ & $2000$  \\
         \hline
         Average number of connections per node & $\langle d \rangle$ & $3$\\
         \hline
         Spectral radius of the reservoir's adjacency matrix & $\rho_s$ & $0.5$\\
         \hline
         Maximal coupling strength of the input to the reservoir & $\chi$ & $0.5$\\
         \hline
         Reservoir leakage parameter & $\alpha$ & $1$\\
         \hline
         Reservoir node activation bias & $b$ & $0$\\
         \hline
         Tikhonov regularization parameter & $\alpha$ & $3.59\times 10^{-13}$\\
         \hline
         Intercept of the linear control signal & $b$ & $1$\\
         \hline
         Slope of the linear control signal & $a$ & $5.99\times 10^{-5}$\\
         \hline
         Numerical integration time step for Eq. (\ref{eqn:ks}) & $\Delta t$ & $0.01$\\
         \hline
         RC time step & $\Delta t_{RC}$ & $0.01$\\
         \hline
         Strength of observational noise added to training data & $\epsilon_0$ & $5\times 10^{-5}$\\
         \hline
         Number of passes of training data during training & &$10$\\
    \end{tabular}
    \label{tab:lorenz_mvt_hp}
\end{table}

\section*{Appendix B: Reservoir Computer Setup for the Ikeda Map Example, Sec. 3.2}

To enable accurate climate prediction in the Ikeda Map example (Sec. 3.2), we modified the reservoir computing architecture in the following two ways. First, we separated the input coupling of the dynamical variables (i.e., $x_n$ and $y_n$) and the linear control signal. More specifically, we coupled the dynamical variables of the system to the reservoir in the same way as before using an input coupling strength $\chi_1$, but the linear control signal was fed to each reservoir node with a $50\%$ probability using an input coupling strength $\chi_2$ which may be different from $\chi_1$. Second, we included the squares of the reservoir node activation in the output  feature vector. With these modifications, the reservoir update function (previously Eq. (\ref{eqn:res_eq})), and the one-step prediction equation become

\begin{subequations}
\begin{equation}
    \label{eqn:res_eq_update}
    \bm{r}(t+\Delta t) = (1-\alpha)\bm{r}(t) + 
    \alpha\tanh{(A\bm{r}(t) + W_{in}^{(1)}\bm{v}(t) + W_{in}^{(2)}s(t) + b_r\mathbb{1}_{N\times 1})},
\end{equation}
\begin{equation}
    \label{eqn:res_output_update}
    \tilde{\bm{v}}(t + \Delta t) = W_{out}[\bm{r}(t + \Delta t);\bm{r}^2(t + \Delta t);\bm{u}(t);1]
\end{equation}
\end{subequations}

where $W_{in}^{(1)}$ is an input-to-reservoir coupling matrix for coupling the dynamical variables $\bm{v}(t)$ to the reservoir and is constructed in the usual way (i.e., by randomly selecting one element from each row to be a non-zero number chosen from a uniform distribution over $[-\chi_1,\chi_1]$), $W_{in}^{(2)}$ is a matrix of size ($N\times 1$) ($N$ is the number of reservoir nodes) to couple the linear control signal $s(t)$ to the reservoir and is constructed by choosing to set each element to be a non-zero number randomly chosen from a uniform distribution over $[-\chi_2,\chi_2]$ with a $50\%$ probability, and $\bm{r}^2$ denotes an element-wise squaring operation. With these modifications we were able to obtain the good climate predictions for the interior crisis in the Ikeda map shown in this section. (Note, we did not find it necessary to make these modifications, and therefore introduce the additional hyperparameter $\chi_2$ or increase the training cost by expanding the feature vector, when considering Lorenz system in the previous section.)

\section*{Appendix C: Hybrid Reservoir Computer Setup and training for the Lorenz system, Sec. 3.4}

\begin{figure}[h]
    \centering
    \includegraphics[scale=0.45]{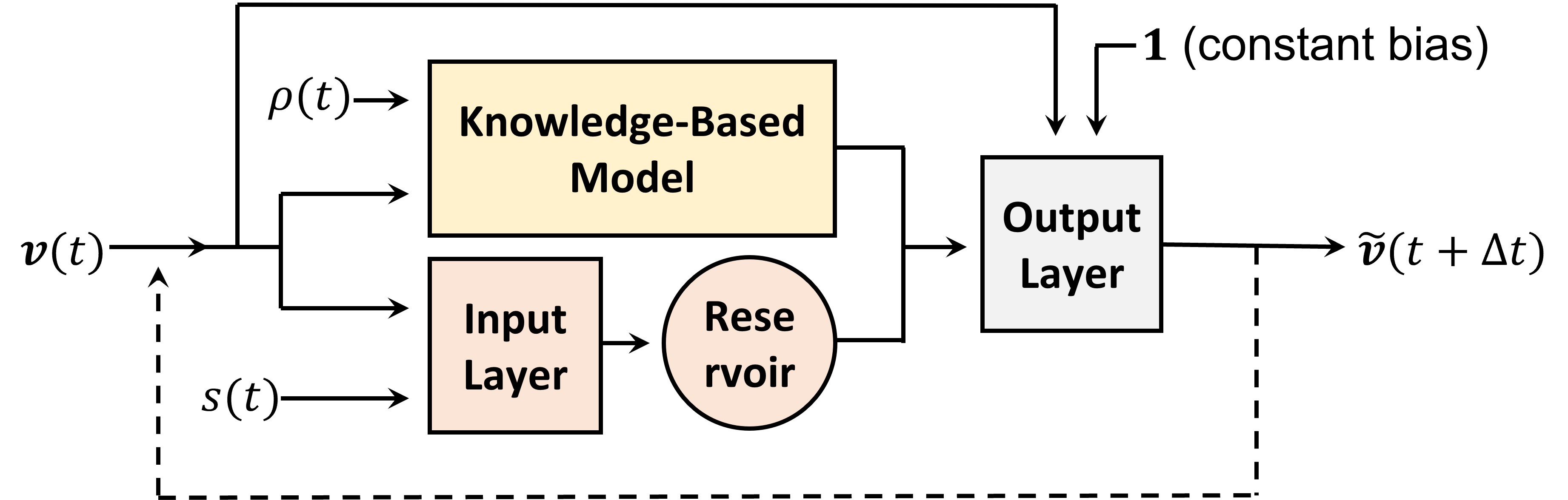}
    \caption{Data-driven-knowledge-based hybrid ML model architecture.}
    \label{fig:hybrid_rc_setup}
\end{figure}

To enable prediction of the hysteretic-bifurcation-mediated tipping point and post-tipping-point dynamics of the Lorenz system in Sec. 3.4, we used a hybrid ML model \cite{pathak_chaos1} which combined a reservoir component driven by the time series data of measured past system states, and a knowledge-based physics model. The architecture is shown in Fig. \ref{fig:hybrid_rc_setup}. Similar to the pure ML model, this hybrid model is operated in an ``open-loop" configuration during training and in a ``closed-loop" configuration during prediction. In the open-loop configuration, observations $\bm{v}(t)$ (for $t \in [-t_t, 0]$) are (1) propagated through the input layer ($W_{in}$) and reservoir to generate the reservoir hidden states $\bm{r}(t+\Delta t)$ according to Eq. (\ref{eqn:res_eq}), and (2) used to produce one-step predictions $\bm{\tilde{v}}_{KB}(t+\Delta t)$ from the knowledge-based model. Feature vectors are constructed using these quantities as $\bm{\tilde{r}}(t + \Delta t) = [\bm{r}(t+\Delta t); \bm{v}(t); 1; \bm{\tilde{v}}_{KB}(t+\Delta t)]$. Training of the output layer is performed by minimizing the following cost function over the training data, 

\begin{equation}
    \frac{1}{t_t} \sum_{-t_t\leq t \leq 0} ||W_{out}\bm{\tilde{r}}(t) - \bm{v}(t) ||^2 + \lambda||W_{out} - P||^2,
    \label{eqn:res_lsq_prior}
\end{equation}

where $P = [\bm{0}\ \mathbb{1}]$ (where $\bm{0}$ is a $(L \times (N+K+1))$ matrix of zeros and $\mathbb{1}$ is a $(L \times L)$ identity matrix). The matrix $P$ represents our prior belief that the knowledge-based model may generate reasonable outputs, from which the ML model outputs should not differ significantly. The closed-loop configuration is used during the prediction phase, where the ML output at time $t$ is recycled as input to both the knowledge-based model component and reservoir component at time $t+\Delta t$. 

\section*{Acknowledgements}

We thank Brian Hunt for insightful comments, suggestions, and advice. The authors acknowledge the University of Maryland supercomputing resources (http://hpcc.umd.edu) made available for conducting the research reported in this paper. This work was supported by DARPA Grant 5298082.

\bibliographystyle{unsrt}
\bibliography{Manuscript}

\begin{thebibliography}{10}

\bibitem{tel_tipping}
B.~Kaszas, U.~Feudel, and T.~Tel.
\newblock Tipping phenomena in typical dynamical systems subjected to parameter
  drift.
\newblock {\em Sci Rep}, 9:8654, 2019.

\bibitem{feudel}
U.~Feudel, A.~N. Pisarchik, and K.~Showalter.
\newblock Multistability and tipping: From mathematics and physics to climate
  and brain - minireview and preface to the focus issue.
\newblock {\em Chaos}, 28:033501, 2018.

\bibitem{jaeger_science}
H.~Jaeger and H.~Haas.
\newblock Harnessing nonlinearity: Predicting chaotic systems and saving energy
  in wireless communication.
\newblock {\em Science}, 304:78, 2004.

\bibitem{pathak_prl}
J.~Pathak, B.~R. Hunt, M.~Girvan, Z.~Lu, and E.~Ott.
\newblock Model-free prediction of large spatiotemporally chaotic systems from
  data: A reservoir computing approach.
\newblock {\em Phys. Rev. Lett.}, 120:024102, 2018.

\bibitem{pathak_chaos1}
J.~Pathak, A.~Wikner, R.~Fussel, S.~Chandra, B.~Hunt, M.~Girvan, and E.~Ott.
\newblock Hybrid forecasting of chaotic processes: Using machine learning in
  conjunction with a knowledge based model.
\newblock {\em Chaos}, 28:041101, 2018.

\bibitem{vaughan}
A.~Vaughan and S.~V. Bohac.
\newblock Real-time, adaptive machine learning for non-stationary, near chaotic
  gasoline engine combustion time series.
\newblock {\em Neural Networks}, 70:18--26, 2015.

\bibitem{griffith}
A.~Griffith, A.~Pomerance, and D.~J. Gauthier.
\newblock Forecasting chaotic systems with very low connectivity reservoir
  computers.
\newblock {\em Chaos}, 29:123108, 2019.

\bibitem{shen}
G.~Shen, J.~Kurths, and Y.~Yuan.
\newblock Sequence-to-sequence prediction of spatiotemporal systems.
\newblock {\em Chaos}, 30:023102, 2020.

\bibitem{follmann}
R.~Follmann and Jr~E.~Rosa.
\newblock Predicting slow and fast neuronal dynamics with machine learning.
\newblock {\em Chaos}, 29:113119, 2019.

\bibitem{herzog}
S.~Herzog, F.~Wörgötter, and U.~Parlitz.
\newblock Convolutional autoencoder and conditional random fields hybrid for
  predicting spatial-temporal chaos.
\newblock {\em Chaos}, 29:123116, 2019.

\bibitem{chatto}
A.~Chattopadhyay, P.~Hassanzadeh, and D.~Subramanian.
\newblock Data-driven predictions of a multiscale {L}orenz 96 chaotic system
  using machine-learning methods: reservoir computing, artificial neural
  network, and long short-term memory network.
\newblock {\em Nonlin. Processes Geophys.}, 27:373--389, 2020.

\bibitem{chen}
D.~Chen and W.~Han.
\newblock Prediction of multivariate chaotic time series via radial basis
  function neural networks.
\newblock {\em Complexity}, 18:55, 2013.

\bibitem{huang}
W.~Huang, Y.~Li, and Y.~Huang.
\newblock Deep hybrid neural network and improved differential neuroevolution
  for chaotic time series prediction.
\newblock {\em IEEE Access}, 8:159552--159565, 2020.

\bibitem{fan}
H.~Fan, J.~Jiang, C.~Zhang, X.~Wang, and Y-C Lai.
\newblock Long-term prediction of chaotic systems with machine learning.
\newblock {\em Phys. Rev. Research}, 2:012080, 2020.

\bibitem{faqih}
Akhmad Faqih, Aldo~Pratama Lianto, and Benyamin Kusumoputro.
\newblock Mackey-{G}lass chaotic time series prediction using modified {RBF}
  neural networks.
\newblock In {\em Proceedings of the 2nd International Conference on Software
  Engineering and Information Management (ICSIM 2019)}, ICSIM 2019, page
  7–11, New York, NY, USA, 2019. Association for Computing Machinery.

\bibitem{gauthier}
Daniel Canaday, Aaron Griffith, and Daniel~J. Gauthier.
\newblock Rapid time series prediction with a hardware-based reservoir
  computer.
\newblock {\em Chaos: An Interdisciplinary Journal of Nonlinear Science},
  28(12):123119, 2018.

\bibitem{borra}
Francesco Borra, Angelo Vulpiani, and Massimo Cencini.
\newblock Effective models and predictability of chaotic multiscale systems via
  machine learning.
\newblock {\em Phys. Rev. E}, 102:052203, Nov 2020.

\bibitem{pathak_chaos2}
J.~Pathak, Z.~Lu, B.~R. Hunt, M.~Girvan, and E.~Ott.
\newblock Using machine learning to replicate chaotic attractors and calculate
  {L}yapunov exponents from data.
\newblock {\em Chaos}, 27:121102, 2017.

\bibitem{lu_chaos}
Z.~Lu, B.~R. Hunt, and E.~Ott.
\newblock Attractor reconstruction by machine learning.
\newblock {\em Chaos}, 28:061104, 2018.

\bibitem{antonik}
P.~Antonik, M.~Gulina, J.~Pauwels, and S.~Massar.
\newblock Using a reservoir computer to learn chaotic attractors, with
  applications in chaos synchronization and cryptography.
\newblock {\em Phys. Rev. E}, 98:012215, 2018.

\bibitem{nguyen}
Duong Nguyen, Said Ouala, Lucas Drumetz, and Ronan Fablet.
\newblock {EM}-like learning chaotic dynamics from noisy and partial
  observations.
\newblock {\em CoRR}, abs/1903.10335, 2019.

\bibitem{ruelle}
J.~P. Eckmann and D.~Ruelle.
\newblock Ergodic theory of chaos and strange attractors.
\newblock {\em Rev. Mod. Phys.}, 57:617, 1985.

\bibitem{drotos1}
G\'abor Dr\'otos, Tam\'as B\'odai, and Tam\'as T\'el.
\newblock Quantifying nonergodicity in nonautonomous dissipative dynamical
  systems: An application to climate change.
\newblock {\em Phys. Rev. E}, 94:022214, Aug 2016.

\bibitem{drotos2}
Gábor Drótos, Tamás Bódai, and Tamás Tél.
\newblock Probabilistic concepts in a changing climate: A snapshot attractor
  picture.
\newblock {\em Journal of Climate}, 28(8), 15 Apr. 2015.

\bibitem{romeiras}
Filipe~J. Romeiras, Celso Grebogi, and Edward Ott.
\newblock Multifractal properties of snapshot attractors of random maps.
\newblock {\em Phys. Rev. A}, 41:784--799, Jan 1990.

\bibitem{chekroun}
Mickaël~D. Chekroun, Eric Simonnet, and Michael Ghil.
\newblock Stochastic climate dynamics: Random attractors and time-dependent
  invariant measures.
\newblock {\em Physica D: Nonlinear Phenomena}, 240(21):1685 -- 1700, 2011.

\bibitem{patel}
D.~Patel, D.~Canaday, M.~Girvan, A.~Pomerance, and E.~Ott.
\newblock Using machine learning to predict statistical properties of
  non-stationary dynamical processes: System climate, regime transitions, and
  the effect of stochasticity.
\newblock {\em Chaos}, 31:033149, 2021.

\bibitem{lim}
Soon~Hoe Lim, Ludovico Theo~Giorgini, Woosok Moon, and J.~S. Wettlaufer.
\newblock Predicting critical transitions in multiscale dynamical systems using
  reservoir computing.
\newblock {\em Chaos: An Interdisciplinary Journal of Nonlinear Science},
  30(12):123126, 2020.

\bibitem{xing}
F.~Z. {Xing}, E.~{Cambria}, and X.~{Zou}.
\newblock Predicting evolving chaotic time series with fuzzy neural networks.
\newblock In {\em 2017 International Joint Conference on Neural Networks
  (IJCNN)}, pages 3176--3183, 2017.

\bibitem{shi}
Zhigang Shi, Yuting Bai, Xuebo Jin, Xiaoyi Wang, Tingli Su, and Jianlei Kong.
\newblock Parallel deep prediction with covariance intersection fusion on
  non-stationary time series.
\newblock {\em Knowledge-Based Systems}, 211:106523, 2021.

\bibitem{pershin}
Anton Pershin, C{\'{e}}dric Beaume, Kuan Li, and Steven~M. Tobias.
\newblock Can neural networks predict dynamics they have never seen?
\newblock {\em CoRR}, abs/2111.06783, 2021.

\bibitem{bury}
Thomas~M. Bury, R.~I. Sujith, Induja Pavithran, Marten Scheffer, Timothy~M.
  Lenton, Madhur Anand, and Chris~T. Bauch.
\newblock Deep learning for early warning signals of tipping points.
\newblock {\em Proceedings of the National Academy of Sciences}, 118(39), 2021.

\bibitem{kong}
Ling-Wei Kong, Hua-Wei Fan, Celso Grebogi, and Ying-Cheng Lai.
\newblock Machine learning prediction of critical transition and system
  collapse.
\newblock {\em Phys. Rev. Research}, 3:013090, Jan 2021.

\bibitem{xiao}
Rui Xiao, Ling-Wei Kong, Zhong-Kui Sun, and Ying-Cheng Lai.
\newblock Predicting amplitude death with machine learning.
\newblock {\em Phys. Rev. E}, 104:014205, Jul 2021.

\bibitem{ghil_review}
M.~Ghil and V.~Lucarini.
\newblock The physics of climate variability and climate change.
\newblock {\em Rev. Mod. Phys.}, 92:035002, 2020.

\bibitem{dekker}
M.~M. Dekker, A.~S. von~der Heydt, and H.~A. Dijkstra.
\newblock Cascading transitions in the climate system.
\newblock {\em Earth Syst. Dynam.}, 9:1243–1260, 2018.

\bibitem{tantet}
A.~Tantet, V.~Lucarini, F.~Lunkeit, and H.~Dijkstra.
\newblock Crisis of the chaotic attractor of a climate model: a transfer
  operator approach.
\newblock {\em Nonlinearity}, 31(5):2221, 2018.

\bibitem{zelnik}
Y.~R. Zelnik, S.~Kinast, H.~Yizhaq, G.~Bel, and E.~Meron.
\newblock Regime shifts in models of dryland vegetation.
\newblock {\em Phil. Trans. R. Soc. A.}, 371:20120358, 2013.

\bibitem{amemiya}
Takashi Amemiya, Takatoshi Enomoto, Axel~G. Rossberg, Tetsuya Yamamoto, Yuhei
  Inamori, and Kiminori Itoh.
\newblock Stability and dynamical behavior in a lake-model and implications for
  regime shifts in real lakes.
\newblock {\em Ecological Modelling}, 206(1):54 -- 62, 2007.

\bibitem{fussmann}
G.~Fussmann, S.~P. Ellner, K.~W. Shertzer, and N.~G.~Hairston Jr.
\newblock Crossing the {H}opf bifurcation in a live predator-prey system.
\newblock {\em Science}, 290:1358--1360, 2000.

\bibitem{alonso}
David Alonso, Andy Dobson, and Mercedes Pascual.
\newblock Critical transitions in malaria transmission models are consistently
  generated by superinfection.
\newblock {\em Philosophical Transactions of the Royal Society B: Biological
  Sciences}, 374(1775):20180275, 2019.

\bibitem{muk}
Zindoga Mukandavire, Shu Liao, Jin Wang, Holly Gaff, David~L. Smith, and
  J.~Glenn Morris.
\newblock Estimating the reproductive numbers for the 2008{\textendash}2009
  cholera outbreaks in {Z}imbabwe.
\newblock {\em Proceedings of the National Academy of Sciences},
  108(21):8767--8772, 2011.

\bibitem{pomeau}
Y.~Pomeau and P.~M. Manneville.
\newblock Intermittent transition to turbulence in dissipative dynamical
  systems.
\newblock {\em Commun. Math. Phys.}, 74:189--197, 1980.

\bibitem{grebogi}
C.~Grebogi, E.~Ott, and J.~A. Yorke.
\newblock Critical exponents of chaotic transients in nonlinear dynamical
  systems.
\newblock {\em Phys. Rev. Lett.}, 57:1284, 1986.

\bibitem{chiriac}
S.~Chiriac, D.~G. Dimitriu, and M.~Sanduloviciu.
\newblock Type {I} intermittency related to the spatiotemporal dynamics of
  double layers and ion-acoustic instabilities in plasma.
\newblock {\em Physics of Plasmas}, 14(7):072309, 2007.

\bibitem{cheung}
P.~Y. Cheung, S.~Donovan, and A.~Y. Wong.
\newblock Observations of intermittent chaos in plasmas.
\newblock {\em Phys. Rev. Lett.}, 61:1360--1363, Sep 1988.

\bibitem{cassak}
P.~A. Cassak, M.~A. Shay, and J.~F. Drake.
\newblock A saddle-node bifurcation model of magnetic reconnection onset.
\newblock {\em Physics of Plasmas}, 17(6):062105, 2010.

\bibitem{chian}
A.~C.-L. Chian, W.~M. Santana, E.~L. Rempel, F.~A. Borotto, T.~Hada, and
  Y.~Kamide.
\newblock Chaos in driven {A}lfvén systems: unstable periodic orbits and
  chaotic saddles.
\newblock {\em Nonlinear Processes in Geophysics}, 14(1):17--29, 2007.

\bibitem{manffra}
Elisangela~Ferretti Manffra, Iber{\^e}~L. Caldas, Ricardo~L. Viana, and
  Hypolito~Jos{\'e} Kalinowski.
\newblock Type-{I} intermittency and crisis-induced intermittency in a
  semiconductor laser under injection current modulation.
\newblock {\em Nonlinear Dynamics}, 27:185--195, 2002.

\bibitem{sanmartin}
J.~San-Martín and J.C. Antoranz.
\newblock {Transition to Chaos via Type-II Intermittency in a Laser with
  Saturable Absorber Externally Excited}.
\newblock {\em Progress of Theoretical Physics}, 94(4):535--542, 10 1995.

\bibitem{dobson}
I.~Dobson.
\newblock Observations on the geometry of saddle node bifurcation and voltage
  collapse in electrical power systems.
\newblock {\em IEEE Transactions on Circuits and Systems I: Fundamental Theory
  and Applications}, 39(3):240--243, 1992.

\bibitem{kim}
Chil-Mim Kim, Geo-Su Yim, Yeon~Soo Kim, Jeong-Moog Kim, and H.~W. Lee.
\newblock Experimental evidence of characteristic relations of type-i
  intermittency in an electronic circuit.
\newblock {\em Phys. Rev. E}, 56:2573--2577, Sep 1997.

\bibitem{huang2}
Jung-Yun Huang and Jong-Jean Kim.
\newblock Type-{II} intermittency in a coupled nonlinear oscillator:
  Experimental observation.
\newblock {\em Phys. Rev. A}, 36:1495--1497, Aug 1987.

\bibitem{guerrero}
L.~E. Guerrero and M.~Octavio.
\newblock Spatiotemporal effects in long rf-biased {J}osephson junctions:
  Chaotic transitions and intermittencies between dynamical attractors.
\newblock {\em Phys. Rev. A}, 40:3371--3380, Sep 1989.

\bibitem{penelet}
Guillaume Penelet, Takumaru Watanabe, and Tetsushi Biwa.
\newblock Study of a thermoacoustic-{S}tirling engine connected to a
  piston-crank-flywheel assembly.
\newblock {\em The Journal of the Acoustical Society of America},
  149(3):1674--1684, 2021.

\bibitem{guan}
Yu~Guan, Vikrant Gupta, and Larry K.~B. Li.
\newblock Intermittency route to self-excited chaotic thermoacoustic
  oscillations.
\newblock {\em Journal of Fluid Mechanics}, 894:R3, 2020.

\bibitem{kabiraj}
Lipika Kabiraj and R.~I. Sujith.
\newblock Nonlinear self-excited thermoacoustic oscillations: intermittency and
  flame blowout.
\newblock {\em Journal of Fluid Mechanics}, 713:376–397, 2012.

\bibitem{ringuet}
E.~Ringuet, C.~Roz\'e, and G.~Gouesbet.
\newblock Experimental observation of type-{II} intermittency in a hydrodynamic
  system.
\newblock {\em Phys. Rev. E}, 47:1405--1407, Feb 1993.

\bibitem{suresha}
Suhas Suresha, R.~I. Sujith, Benjamin Emerson, and Tim Lieuwen.
\newblock Nonlinear dynamics and intermittency in a turbulent reacting wake
  with density ratio as bifurcation parameter.
\newblock {\em Phys. Rev. E}, 94:042206, Oct 2016.

\bibitem{krischer}
K.~Krischer, M.~Lübke, W.~Wolf, M.~Eiswirth, and G.~Ertl.
\newblock Chaos and interior crisis in an electrochemical reaction.
\newblock {\em Berichte der Bunsengesellschaft für physikalische Chemie},
  95(7):820--823, 1991.

\bibitem{luk}
M.~Lukosevicius and H.~Jaeger.
\newblock Reservoir computing approaches to recurrent neural network training.
\newblock {\em Comput. Sci. Rev.}, 3:127–149, 2009.

\bibitem{jaeger2}
H.~Jaeger.
\newblock The “echo state” approach to analyzing and training recurrent
  neural networks-with an erratum note.
\newblock {GMD Technical Report No.} 148, German National Research Center for
  Information Technology, Bonn, Germany, 2001.

\bibitem{maass}
W.~Maass, T.~Natschlager, and H.~Markram.
\newblock Real-time computing without stable states: A new framework for neural
  computation based on perturbations.
\newblock {\em Neural Comput.}, 14:2531–2560, 2002.

\bibitem{lorenz}
E.~Lorenz.
\newblock Deterministic nonperiodic flow.
\newblock {\em Journal of Atmospheric Sciences}, 20:130, 1963.

\bibitem{hammel}
S.~M. Hammel, C.~K. R.~T. Jones, and J.~V. Moloney.
\newblock Global dynamical behavior of the optical field in a ring cavity.
\newblock {\em J. Opt. Soc. Am. B}, 2(4):552--564, Apr 1985.

\bibitem{kuramoto}
Y.~Kuramoto.
\newblock Diffusion-induced chaos in reaction systems.
\newblock {\em Progress of Theoretical Physics Supplement}, 64:346--367, 1978.

\bibitem{siva}
G.~I. Sivashinsky.
\newblock On flame propagation under conditions of stoichiometry.
\newblock {\em SIAM J. Appl. Math.}, 39:67--82, 1980.

\bibitem{chandra}
S.~Chandra, B.~R. Hunt, R.~Roy, and E.~Ott.
\newblock To be published.

\bibitem{wikner}
A.~Wikner, B.~Hunt, J.~Harvey, M.~Girvan, and E.~Ott.
\newblock To be published.

\bibitem{vallender}
S.~S. Vallender.
\newblock Calculation of the {W}asserstein distance between probability
  distributions on the line.
\newblock {\em Theory Probab. Appl.}, 18:784--786, 1973.

\bibitem{mishra}
Arindam Mishra, S.~Leo~Kingston, Chittaranjan Hens, Tomasz Kapitaniak, Ulrike
  Feudel, and Syamal~K. Dana.
\newblock Routes to extreme events in dynamical systems: Dynamical and
  statistical characteristics.
\newblock {\em Chaos: An Interdisciplinary Journal of Nonlinear Science},
  30(6):063114, 2020.

\bibitem{trenberth}
Kevin~E. Trenberth, Aiguo Dai, Roy~M. Rasmussen, and David~B. Parsons.
\newblock The changing character of precipitation.
\newblock {\em Bulletin of the American Meteorological Society}, 84(9):1205 --
  1218, 2003.

\bibitem{goswami}
B.~N. Goswami, V.~Venugopal, D.~Sengupta, M.~S. Madhusoodanan, and Prince~K.
  Xavier.
\newblock Increasing trend of extreme rain events over india in a warming
  environment.
\newblock {\em Science}, 314(5804):1442--1445, 2006.

\bibitem{ray}
Arnob Ray, Sarbendu Rakshit, Dibakar Ghosh, and Syamal~K. Dana.
\newblock Intermittent large deviation of chaotic trajectory in the {I}keda
  map: Signature of extreme events.
\newblock {\em Chaos: An Interdisciplinary Journal of Nonlinear Science},
  29(4):043131, 2019.

\bibitem{kassam}
Aly-Khan Kassam and Lloyd~N. Trefethen.
\newblock Fourth-order time-stepping for stiff {PDE}s.
\newblock {\em SIAM Journal on Scientific Computing}, 26(4):1214--1233, 2005.

\bibitem{edson}
Russell~A. Edson, J.~E. Bunder, Trent~W. Mattner, and A.~J. Roberts.
\newblock Lyapunov exponents of the {K}uramoto–{S}ivashinsky {PDE}.
\newblock {\em The ANZIAM Journal}, 61(3):270–285, 2019.

\bibitem{yorke}
J.~A. Yorke and E.~D. Yorke.
\newblock Metastable chaos: The transition to sustained chaotic behavior in the
  lorenz model.
\newblock {\em J Stat Phys}, 21:263--277, 1979.

\bibitem{ott}
Edward Ott.
\newblock {\em Chaos in Dynamical Systems}.
\newblock Cambridge University Press, 2 edition, 2002.

\end{thebibliography}
\end{document}